%% file: root.tex
\documentclass[letterpaper, 10 pt, conference]{ieeeconf}  % Comment this line out if you need a4paper

\IEEEoverridecommandlockouts                              % This command is only needed if 
\usepackage{cite}
\usepackage{amsmath}
\usepackage{url}
\usepackage{outlines}

\makeatletter
\let\NAT@parse\undefined
\makeatother
\usepackage[bookmarks=true]{hyperref}
\hypersetup{
colorlinks=true,
urlcolor=blue,
}
\usepackage[footnotesize]{caption}

\usepackage{svg}
\usepackage{cleveref}

\usepackage{amsfonts}
\usepackage{tensor}
\usepackage{bm}

\usepackage{siunitx}
\usepackage{gensymb}
\usepackage{xargs}
\usepackage{subcaption}
\usepackage{hhline}
\usepackage{multirow}

\usepackage{listings}

\usepackage{placeins} % for \FloatBarrier
\usepackage{makecell}
\usepackage{comment}

\includecomment{appendixcond} % Adds Appendix
\input{math_commands}

\newcommand{\sturmauthor}{Sturm et al.~\cite{sturm_probabilistic_2011} }

\newcommand{\screwnetname}{%
ScrewNet~\cite{jain_screwnet_2021} }

\title{\LARGE \bf 
\textbf{C}ategory-Independent \textbf{A}rticulated \textbf{O}bject \textbf{T}racking with \textbf{F}actor \textbf{G}raphs
}

\author{Nick~Heppert$^{1}$,~Toki~Migimatsu$^{2}$,~Brent~Yi$^{3}$,~Claire~Chen$^{2}$,~and~Jeannette~Bohg$^{2}$% <-this % stops a space
\thanks{$^{1}$Nick Heppert was with the Department of Computer Science, TU Darmstadt, Germany while doing this work {\tt\small heppert@cs.uni-freiburg.de}}% <-this % stops a space
\thanks{$^{2}$Toki Migimatsu, Claire Chen, and Jeannette Bohg are with the Department of Computer Science, Stanford University, USA  {\tt\small \{takatoki,clairech,bohg\}@stanford.edu}}
\thanks{$^{3}$Brent Yi is with the Department of Electrical Engineering and Computer Sciences, UC Berkeley, USA {\tt\small brentyi@berkeley.edu}}% <-this % stops a space
\thanks{The authors would like to thank Jan Peters from TU Darmstadt, Germany for discussion and Yijia Weng from Stanford University, USA for providing evaluation test predictions from CAPTRA.}% <-this % stops a space
}

\begin{document}

\maketitle
\thispagestyle{empty}
\pagestyle{empty}

\begin{abstract}
\input{tex/0_abstract.tex}
\end{abstract}

% \begin{IEEEkeywords}
% IEEE, IEEEtran, journal, \LaTeX, paper, template.
% \end{IEEEkeywords}

% \fi
\IEEEpeerreviewmaketitle

\section{Introduction}
% for IEEE journal papers produced under \LaTeX\ using
% IEEEtran.cls version 1.8b and later.

\input{tex/1_intro.tex}
\section{Related Work}
\input{tex/2_related.tex}
\section{Background: Twist Joint Representation}
\label{sec:twist_joint_representation}
\input{tex/3_twist.tex}

\section{Category-Independent Part Pose Tracking}
\label{subsec:methodology:category_independent_detection}
\input{tex/4_vision.tex}

\section{Articulation Model Estimation}
\label{sec:factor_graph}
\input{tex/5_fg.tex}

\section{Tangent Similarity Metric}
\label{sec:tangent_sim_metric}
\input{tex/6_metric.tex}

\section{Experiments}
\input{tex/7a_fg_exp.tex}
\input{tex/7b_full_exp.tex}

\section{Conclusion}
\input{tex/8_conclusion.tex}
\section*{Acknowledgements}
\input{tex/9_acknowledgements.tex}


% \newpage
% \Floatbarrier

%\section*{Acknowledgment}


% \IEEEtriggeratref{0}


\bibliographystyle{IEEEtran}
\bibliography{IEEEabrv,msc_thesis}

\begin{appendixcond}
\newpage
\appendices
\section{Tangent Similarity Metric for \textbf{ScrewNet} Predictions}
\label{appendix:sec:metric_screwnet}
\input{tex/A_screwnet.tex}

%\section{Training of Our Visual Perception Module}
%\input{tex/B_vision.tex}

\section{Experiment Data}
\label{appendix:sec:experiment_data}
\input{tex/C_experiment.tex}
\end{appendixcond}

\end{document}

%% file: math_commands.tex
\newcommandx*{\transform}[2][1=i, 2=j]{
	\tensor[^{#1}]{\bm{T}}{_{#2}}
}

\newcommandx*{\poselat}[2][1=i, 2=j]{
	\tensor[^{#1}]{\bm{x}}{_{#2}}
}

\newcommandx*{\poseobs}[2][1=i, 2=j]{
	\tensor[^{#1}]{\bm{y}}{_{#2}}
}

\newcommandx*{\transformobs}[2][1=i, 2=j]{
	\tensor[^{#1}]{\bm{Z}}{_{#2}}
}

\newcommandx*{\transformobshat}[2][1=i, 2=j]{
	\tensor[^{#1}]{\hat{\bm{Z}}}{_{#2}}
}

\newcommand{\realspace}[1]{
	\ensuremath{\mathbb{R}^{#1}}
}

\newcommand{\lieExp}[1]{
	\ensuremath{\operatorname{Exp}\left({#1}\right)}
}

\renewcommand{\vec}[1]{\bm{#1}}

\newcommand{\crossm}[1]{[#1]_\times}

\DeclareMathOperator*{\argmin}{arg\,min}

%% file: tex/0_abstract.tex
Robots deployed in human-centric environments may need to manipulate a diverse range of articulated objects, such as doors, dishwashers, and cabinets. Articulated objects often come with unexpected articulation mechanisms that are inconsistent with categorical priors: for example, a drawer might rotate about a hinge joint instead of sliding open. We propose a category-independent framework for predicting the articulation models of unknown objects from sequences of RGB-D images. The prediction is performed by a two-step process: first, a visual perception module tracks object part poses from raw images, and second, a factor graph takes these poses and infers the articulation model including the current configuration between the parts as a 6D twist. We also propose a manipulation-oriented metric to evaluate predicted joint twists in terms of how well a compliant robot controller would be able to manipulate the articulated object given the predicted twist. We demonstrate that our visual perception and factor graph modules outperform baselines
% on both simulated and real world data.
% on simulated data and show the applicability of our approach on real world data.
on simulated data and show the applicability of our factor graph on real world data. Videos and the source code are available on our project page \url{https://tinyurl.com/ycyva37v}.

%% file: tex/1_intro.tex
% Robots deployed in human environments often interact with objects such as doorways, storage furniture, and appliances like dishwashers or refrigerators. All these objects contain movable parts essential to their function, such as doors or drawers. %, and often have knobs, buttons, or handles to facilitate operation.
Useful tasks in human environments often require interactions with objects such as doorways, storage furniture, and appliances like dishwashers or refrigerators. All these objects contain movable parts essential to their function, such as doors or drawers. %, and often have knobs, buttons, or handles to facilitate operation.
These types of objects, which have at least two rigid parts connected by movable joints, are known as articulated objects. To manipulate articulated objects, robots must be able to track parts of the object and estimate how they move. While articulated objects are commonplace, their appearance, geometry, and kinematics vary greatly, making it crucial to design tracking and estimation methods that work with as little prior knowledge as possible and are robust to vast degrees of variation.

\begin{figure}
    \centering
    \setlength{\belowcaptionskip}{0mm}  % Space below caption.
    \setlength{\textfloatsep}{2mm}  % Space between figures and text.
    \setlength{\dbltextfloatsep}{2mm}  % Space between figures and text.
    \includegraphics[width=\linewidth]{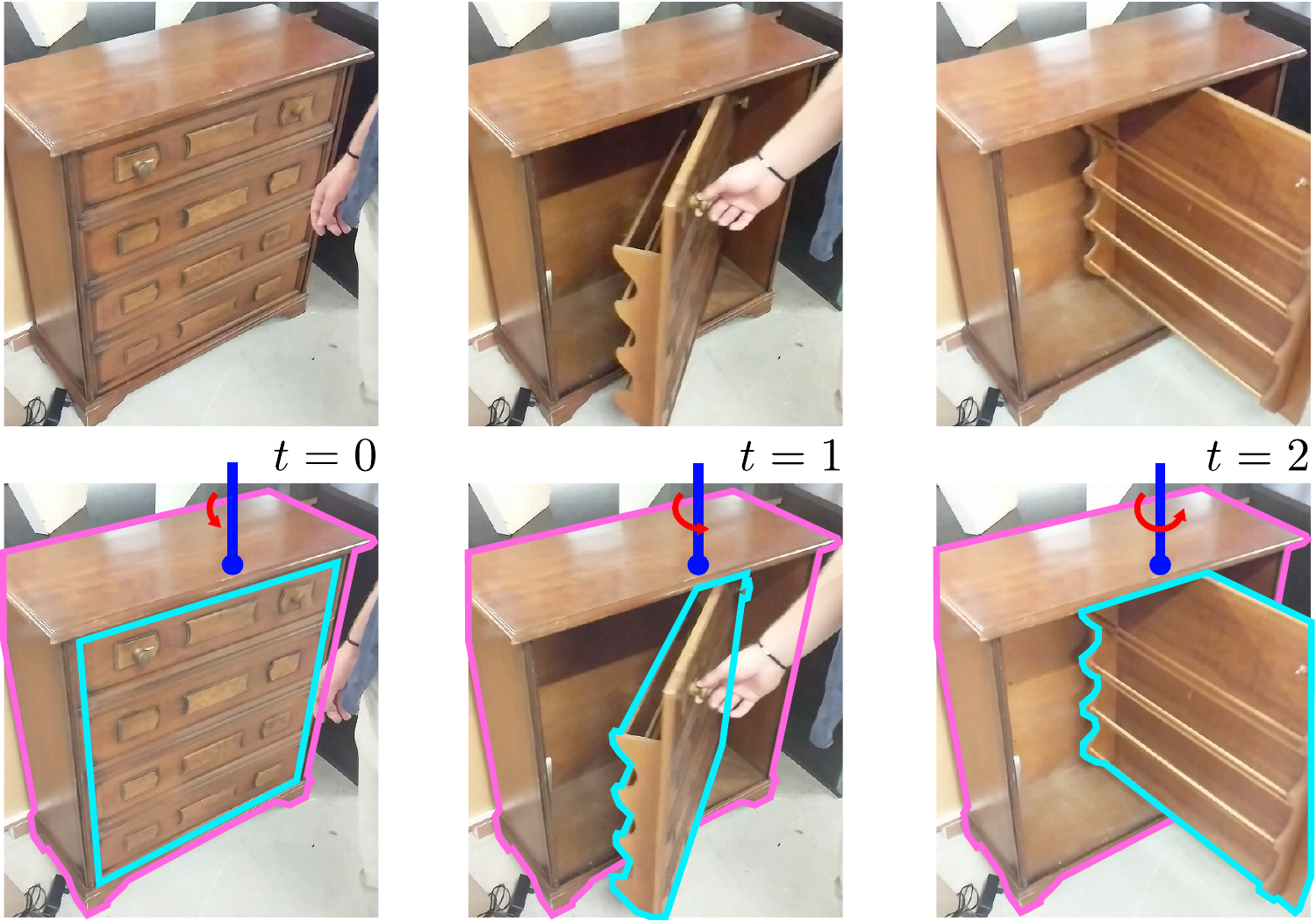}
    \caption{
        Object articulations often contradict categorical priors.
        While one would typically expect this drawer to contain prismatic joints, it actually has a single revolute joint.
        Our system is \textit{category-independent}, predicting part poses (pink, cyan), joint parameters (blue), and time-varying configurations (red) from RGB-D sequences without category bias.
    }
    \vspace{-0.4cm}
    \label{fig:intro}
\end{figure}

We propose a two-step approach to estimate articulation models, consisting of time-invariant joint parameters and time-varying joint states. Our method comprises a category-independent visual perception module for tracking part poses and a factor graph that estimates the articulation model governing part movements. First, we introduce a part pose tracking module that takes a sequence of \mbox{RGB-D} images of an object in motion as input and predicts part poses at every timestep. Second, we formalize articulation model estimation with a novel, factor graph-based structure. Notably, this two-step approach predicts articulation models without a category bias, thus providing generalization to novel instances without the need to first label them with the correct category. This is accomplished by first predicting and tracking part poses with learned motion features that we show are abstract enough to generalize to categories not seen during training, and second, by formulating the joint as a twist, which, compared to prior work \cite{pavlasek_parts-based_2020, martin-martin_coupled_nodate,sturm_probabilistic_2011, jain_learning_2020, amato_multiview_2020, abbatematteo_learning_2019, michel_pose_2015, li_category-level_2020, zeng_visual_2020, DBLP:conf/rss/PillaiWT14, hausman_active_2015}, unifies the representation of prismatic, revolute, and helical joints. Fig.~\ref{fig:intro}  highlights the need for category-independent articulation model prediction.

Finally, we propose a manipulation-oriented tangent similarity metric for evaluating articulation model estimates. Prior works evaluate the accuracy of the predicted joint types and axes \cite{sturm_probabilistic_2011, abbatematteo_learning_2019}.
However, these metrics do not reflect how successful a robot might be at manipulating the articulated object given the predicted articulation model. Compliant controllers need to know the tangent direction in which to apply a force at a given time step, not necessarily whether the joint is prismatic or revolute \cite{prats_compliant_2010,martin_online_2014}. To that end, our proposed tangent similarity metric measures how accurately an estimated articulation model can predict the tangent direction for manipulation.
%However, compliant controllers, such as those used in \cite{prats_compliant_2010} and \cite{martin_online_2014}, require knowing the direction in which to appmove the part at a given only require knowing a single direction in which the robot should move the part rather than needing a full characterization of the object's articulation model. To that end, our proposed metric captures the accuracy of local information about how an object part moves given the estimated articulation model.
%how well a compliant controller would be able to manipulate the object to evaluate articulation predictions. For example, the compliant controllers used to interact with articulated objects in \cite{prats_compliant_2010} and \cite{martin_online_2014} only require specifying a single local direction in which the robot should move the part. Thus, the aim of our work is to provide downstream compliant controllers with the information required to manipulate articulated objects. To that end, our proposed metric only captures local information about how an object part moves when constrained by a certain joint type.

Using this metric, we first test the factor graph's ability to predict articulation models from noisy pose observations and show that is more robust than a well-known baseline~\cite{sturm_probabilistic_2011}. Second, we show that our full tracking and estimation pipeline outperforms another category-independent method~\cite{jain_screwnet_2021}.

%% file: tex/2_related.tex
% Only arxiv: DITTO  add?

% What about OMAD: Object Model with Articulated Deformations for Pose Estimation and Retrieval?

% moses_visual_2020 removed, not close enough to our work

%In the following, we review existing work on tracking and estimation of articulated objects along different dimensions.
%In the following, we position our work relative to existing work on tracking and estimation of articulated objects along several axes.

% We focus on work that estimates the model of the articulation mechanism based on visual data. \cite{michel_pose_2015, martin-martin_coupled_nodate, li_category-level_2020, moses_visual_2020, abbatematteo_learning_2019, jain_screwnet_2021, zeng_visual_2020, liu_nothing_2020, jain_learning_2020, sturm_probabilistic_2011, hausman_active_2015}
% Similar to our work, some of these also incorporate a part-level tracking \cite{weng_captra_2021, martin-martin_coupled_nodate, pavlasek_parts-based_2020, michel_pose_2015, li_category-level_2020, liu_nothing_2020}.

\subsubsection{Prior Object Knowledge}
\label{subsubsec:related:prior-object-knowledge}
Existing estimation methods can be categorized into three groups based on the scope of tracked objects they can handle. \textit{Instance-level} methods only estimate joint configurations for known or previously seen object instances \cite{pavlasek_parts-based_2020, michel_pose_2015, DBLP:conf/rss/PillaiWT14}.
%First, there is a variety of methods that only estimate joint configurations for \textit{known or previously seen object instances} \cite{pavlasek_parts-based_2020, michel_pose_2015, DBLP:conf/rss/PillaiWT14}.
Increasing in difficulty, \textit{category-level} methods can handle unseen instances within a known category \cite{li_category-level_2020, abbatematteo_learning_2019, weng_captra_2021}. Finally, \textit{category-independent} methods estimate joint mechanisms without any category information \cite{jain_screwnet_2021, zeng_visual_2020, liu_nothing_2020, jain_learning_2020, martin-martin_coupled_nodate, sturm_probabilistic_2011, hausman_active_2015, DBLP:conf/corl/JainGLN21, amato_multiview_2020, yi_deep_2019, DBLP:conf/cvpr/YewL20, wang_shape2motion_2019}. In this paper, we propose a novel category-independent method, which assumes minimal prior knowledge about the object to avoid category bias.

\subsubsection{Input Modality}
% mo_where2act_2021 --> single observation
Many works assume that a method to track part poses already exists and instead focus on articulation model estimation only, where the input is a \textit{sequence of 6D poses} \cite{sturm_probabilistic_2011, hausman_active_2015, jain_learning_2020}. When used as a standalone module, the factor graph portion of our two-part pipeline can be seen as a member of this class.

Other works attempt to predict the articulation mechanism from a single observation, which could be an image \cite{DBLP:conf/rss/PillaiWT14, michel_pose_2015, abbatematteo_learning_2019, pavlasek_parts-based_2020, li_category-level_2020, zeng_visual_2020} or a (partial) point cloud \cite{DBLP:conf/cvpr/YewL20, wang_shape2motion_2019}. \cite{shao_motion-based_2018} propose a general object motion predictor for two consecutive RGB-D frames. Similarly, \cite{yi_deep_2019} use two point clouds with the object in two different states.
However, it has been shown that an observation \textit{sequences} can greatly improve estimates over time \cite{sturm_probabilistic_2011, martin-martin_coupled_nodate}.

To that end, many works track part poses from a sequence of images, either with a segmentation masks \cite{jain_screwnet_2021, DBLP:conf/corl/JainGLN21, zeng_visual_2020} or without \cite{liu_nothing_2020, amato_multiview_2020}.
% Among the works that handle the part tracking problem from a sequence of images \cite{liu_nothing_2020, amato_multiview_2020}, some additionally require segmentation masks \cite{jain_screwnet_2021, DBLP:conf/corl/JainGLN21, zeng_visual_2020}.  
% tracks articulated objects with revolute joints from RGB-D images, but only outputs joint angles instead of part poses.
\cite{weng_captra_2021} track part poses from point cloud observations for known categories. \cite{huang_multibodysync_2021} do not track over time but operate on the full set of point clouds and synchronize all of them.

Closest to our work, \cite{martin-martin_coupled_nodate} tracks part poses from RGB-D images, but the image features they use require visually textured objects. While our part tracking module also tracks part poses from RGB-D images, we do not make assumptions about the visual appearance of objects.
%Existing methods either take only a \textit{single visual observation} as input \cite{li_category-level_2020, zeng_visual_2020, pavlasek_parts-based_2020, abbatematteo_learning_2019, michel_pose_2015, DBLP:conf/rss/PillaiWT14} or they use a \textit{sequence of visual observations} \cite{jain_screwnet_2021, liu_nothing_2020, martin-martin_coupled_nodate, weng_captra_2021, DBLP:conf/corl/JainGLN21, sturm_probabilistic_2011} or \textit{of 6D poses} \cite{sturm_probabilistic_2011, jain_learning_2020}. 

%\cite{amato_multiview_2020} uses a sequence of visual observations combined with an utterance describing the scene. \cite{hausman_active_2015} uses a sequence of 6D poses and fuses them with applied action outcomes.
%
%Our work uses a sequence of observations (either poses or RGB-D images), which has been shown to improve estimates over time \cite{sturm_probabilistic_2011, martin-martin_coupled_nodate}. On the visual perception side, additional priors such as masks that either \textit{jointly segments two parts of interest} \cite{jain_screwnet_2021, DBLP:conf/corl/JainGLN21} or \textit{segments each part by itself} \cite{zeng_visual_2020} could be used. \cite{pavlasek_parts-based_2020} assumes to know the \textit{3D CAD model} of the object allowing to render ad-hoc segmentation masks. Compared to these approaches, our visual perception module does not use any explicit (visual) priors as mentioned and can directly use raw pixels to detect parts. 

\subsubsection{Supported Joint Types}
%There are four joint types that are usually considered when tracking articulated objects. Commonly it is assumed that parts are either rigidly connected or have no connection at all. 
Commonly, it is assumed that parts are connected by a $1$-degree of freedom (DoF) joint---either a \textit{revolute hinge} or a \textit{prismatic slider} \cite{pavlasek_parts-based_2020, martin-martin_coupled_nodate,sturm_probabilistic_2011, jain_learning_2020, amato_multiview_2020, abbatematteo_learning_2019, michel_pose_2015, li_category-level_2020, zeng_visual_2020, DBLP:conf/rss/PillaiWT14, hausman_active_2015, DBLP:conf/cvpr/YewL20}. 
% , weng_captra_2021 % UNSURE makes no assumption about the joint, could be added
%With a \textit{screw representation}, ScrewNet \cite{jain_screwnet_2021,DBLP:conf/corl/JainGLN21} can additionally model $1$-dof helical joints (e.g. bottle screw caps).
\sturmauthor are able to model joint types with up to five degrees of freedom by fitting a \textit{Gaussian process} to part pose observations. In contrast to this non-parametric approach, \cite{rofer_kineverse_2022} proposed a \textit{symbolic modeling language} for arbitrary articulated objects.

% Our supported types: (rigid), revolute, prismatic, helical

In this work, we adopt and modify the \textit{screw representation} of ScrewNet~\cite{jain_screwnet_2021,DBLP:conf/corl/JainGLN21}. While other approaches (e.g. \cite{sturm_probabilistic_2011, martin-martin_coupled_nodate}) require different representations for prismatic/revolute joints and different logic to process each, the screw representation generalizes $1$-DoF prismatic, revolute, and helical joints into a single continuous representation.

\subsubsection{Supported Kinematic Structures}
While the joint type only models a local view of how two parts are connected, we are also interested in the overall kinematic structure of an object. The most general representation of this structure is through a \textit{graph}, which allows arbitrarily connected parts \cite{pavlasek_parts-based_2020, martin-martin_coupled_nodate, abbatematteo_learning_2019, DBLP:conf/rss/PillaiWT14}
%, li_category-level_2020 % UNSURE energy term is optimized for all joint types together --> implicit graph
%, weng_captra_2021 % UNSURE makes no assumption about the structure, could be added
and even closed-loop kinematic chains with interdependence between degrees of freedoms \cite{sturm_probabilistic_2011}. \cite{jain_learning_2020} propose configuration-dependent changes in a graph structure.
% Interdependence between degrees of freedom in a graph is possible, e.g. through loops in the graph~\cite{sturm_probabilistic_2011}.
% So, even though there are more possible degrees of freedom in the graph, it can be mapped to a smaller amount of degrees of freedom \cite{sturm_probabilistic_2011}.
Other works have simplified the general graph structure to a \textit{tree} \cite{amato_multiview_2020} or \textit{chain} \cite{liu_nothing_2020, michel_pose_2015}, or abstained from considering the full kinematic structure by keeping to only local views of \textit{two parts} \cite{jain_screwnet_2021, DBLP:conf/corl/JainGLN21, zeng_visual_2020, hausman_active_2015}.
% These works have no need to represent the kinematic structure in any way.

Our factor graph-based approach is inherently able to represent articulated objects with arbitrary kinematic structures. However, we restrict the experimental evaluation to articulated objects with a single prismatic or revolute joint.

% \subsubsection{Additional Priors}
% While the instance-level and category-level information described in Sec.~\hyperref[subsubsec:related:prior-object-knowledge]{II.1} is also a prior, we want to further differentiate between additional priors.

% % mo_where2act_2021 --> no prior information
% On the structure and state estimation side of the problem there are methods that \textit{do not} need any prior information at all \cite{sturm_probabilistic_2011} (through the non-parametric GP joint model) and just operate on the input data. Related to this, one could consider the set of possible \textit{joint types} as a prior itself and thus interpret works that at the first glance seem to have no prior such as \cite{martin-martin_coupled_nodate} as having a weak prior on the possible objects to track. Making the problem slightly easier, we could additionally provide some \textit{geometric constraints} as in \cite{liu_nothing_2020, michel_pose_2015}. A geometric constraint could be for instance that we know how many parts our object consists of. Our factor graph estimation contribution has the implicit prior of the fixed set of joints (rigid, revolute, prismatic, helic) it can represent.

% \subsection{Fitting our approach in here}
% No previous approach works on a category-free level, representing all 3 joint types (prismatic, revolute and helical joints) and being able to represent arbitrary kinematic structures through a graph structure with no prior knowledge. Given the above dimensions the only way to make this problem harder is to operate on a single image, rather than a sequence.

\subsubsection{Output Representation}
The classical representation for articulation models is a joint axis paired with the joint type \cite{sturm_probabilistic_2011, martin-martin_coupled_nodate, DBLP:conf/rss/PillaiWT14, hausman_active_2015, michel_pose_2015, abbatematteo_learning_2019, li_category-level_2020, jain_learning_2020, pavlasek_parts-based_2020, amato_multiview_2020, zeng_visual_2020, DBLP:conf/cvpr/YewL20}. Like \cite{jain_screwnet_2021,DBLP:conf/corl/JainGLN21}, we output joint twists instead, which avoids complications with optimizing over the hybrid continuous-discrete space of the classical representation. One might attempt to classify the joint type (prismatic vs. revolute) from the predicted twist. However, to successfully manipulate an articulated object, accurately predicting the ground truth joint type is unnecessary; compliant controllers can move articulated parts simply by applying the correct force in the object's local frame \cite{prats_compliant_2010, hausman_active_2015, xu2022umpnet}. Based on this idea, we formalize a new metric for evaluating articulated object models independent of their underlying ground truth kinematics.

%% file: tex/3_twist.tex
% We represent prismatic, revolute, and helical joints with general functions that describe the constrained motion between two rigid bodies connected by a joint. Specifically, we express the pose of a child body relative to its parent as the function \mbox{$f( \bm{q} ; \bm{\Theta} ) \in SE(3)$}, where $\bm{\Theta}$ are the joint parameters and $\bm{q}$ is the time-varying joint configuration. 
% The joint parameters $\bm{\Theta}$ consists of the twist $\bm{\nu} = (\bm{v}, \bm{\omega}) \in se(3)$, where $\bm{v}, \bm{\omega} \in \realspace{3}$ describe the linear and angular motion, respectively. The joint configuration is single scalar $q \in \realspace{}$.

% The helical joint function is defined as
% \begin{align}
%     f_{\text{twist}} \left(q; \bm{\nu} \right)
%         &= \lieExp{q \bm{\nu}} \in SE(3), \label{eqn:twist_joint_formulation}
% \end{align}
% where $\operatorname{Exp}$ is the Lie exponential map.

% Prismatic and revolute joints are special cases of helical joints, where $\bm{\omega} = \bm{0}$ for prismatic joints and $\bm{v} \times \bm{\omega} = \bm{0}$ for revolute joints.

% A joint describes the motion constraint between two rigid bodies. The constrained motion is defined by the
We represent a motion constraint between two connected rigid bodies with
joint parameters $\bm{\Theta}$ and time-varying joint configuration $\bm{q}$. The pose of the child body relative to its parent can be expressed as a function $f( \bm{q} ; \bm{\Theta} ) \in SE(3)$.

Specifically, in the case of helical joints, the joint parameters $\bm{\Theta}$ consists of the twist $\bm{\nu} = (\bm{v}, \bm{\omega}) \in se(3)$, where $\bm{v}, \bm{\omega} \in \realspace{3}$ describe the linear and angular motion, respectively. $se(3)$ is the Lie algebra of $SE(3)$. The joint configuration can be specified as a single scalar $q \in \realspace{}$. The helical joint function is defined as
\begin{align}
    f_{\text{twist}} \left(q; \bm{\nu} \right)
        &= \lieExp{q \bm{\nu}} \in SE(3), \label{eqn:twist_joint_formulation}
\end{align}
where $\operatorname{Exp}$ is the Lie exponential map.

Prismatic and revolute joints are special cases of helical joints, where $\bm{\omega} = \bm{0}$ for prismatic joints and $\bm{v} \times \bm{\omega} = \bm{0}$ for revolute joints.

%% file: tex/4_vision.tex
\begin{figure}
    \centering
    \includegraphics[width=\linewidth]{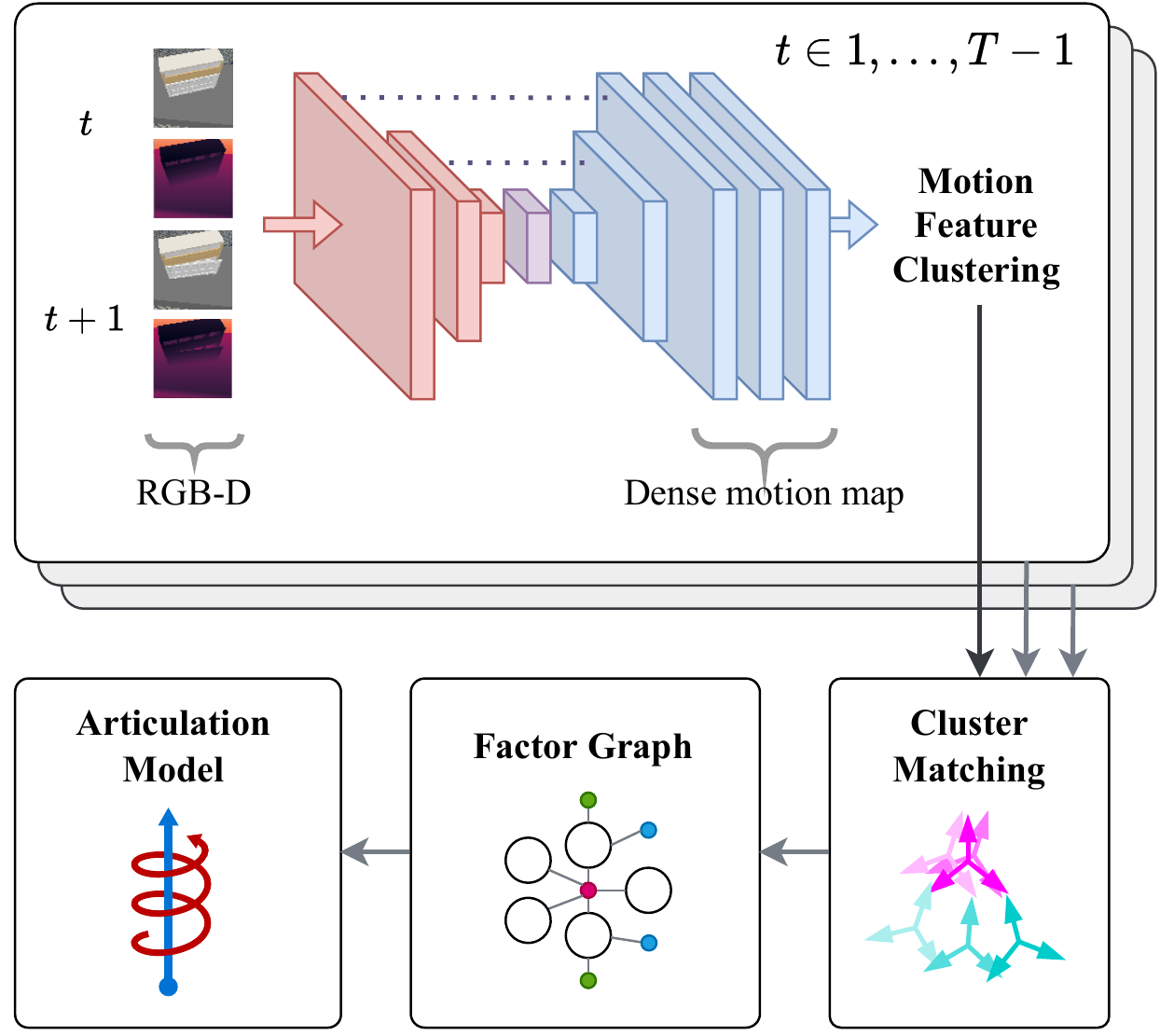}
    \caption{Full articulation model estimation pipeline. To detect and track part poses from RGB-D images, we first feed a pair of images from consecutive time steps to a motion feature predictor. This predictor outputs a pixel-level motion map that represents the center and delta pose of the part in each pixel. The motion map is then segmented into clusters, where the pixels in each cluster are aggregated to output the center and delta pose of a distinct part at the current time step. Detected parts are matched between time steps to form connected trajectories. The factor graph takes these predicted part trajectories and estimates the articulation model.}
    % \\\todo{few minor points: -Let's give the dense motion map either 3 distinct colors (for each quantity one) or drop the green ones?; Something I did wrong previously, there are  1, ..., T-1 *commas* missing for the time variable?; Remove joint type and state and replace with articulation mechanism/model as we do not mention it all in the text?; Make the red arrow helical to show general twist capability?}}
    \label{fig:full_pipeline}
\end{figure}

Our method is a two-step pipeline, shown in Fig.~\ref{fig:full_pipeline}, comprised of a category-independent part pose tracking module and a factor graph-based articulation estimation module. The part pose tracking module, described in this section, detects and tracks poses of an articulated object's parts from \mbox{RGB-D} images without any prior knowledge about the object's category. The factor graph estimation module, described in Sec.~\ref{sec:factor_graph}, takes in these part trajectories to estimate the articulation model.

The part pose tracking module takes in $T$ consecutive RGB-D images of a scene with one articulated object and outputs $T$ poses (part center and delta pose) for each part. We do not assume that the object or parts are segmented. The poses are represented in the camera frame.

The tracking is broken down into three steps: 1) a feature predictor encodes a pair of consecutive RGB-D images into a dense motion map, 2) the dense motion map is segmented into clusters representing detected parts, where the mean of the features within a cluster represents the predicted motion for the corresponding part at the current time step, and 3) the detected parts for the current time step are matched and appended to trajectory predictions from previous time steps. The sub-sections below describe each of these steps in detail.

% Lastly, we introduce our new visual perception module that is able to detect and track parts (of an articulated object) on a sequence of RGB-XYZ images without any prior knowledge required.
% The full pipeline is given in \Cref{fig:full_pipeline}. In brief, for each pair of consecutive time steps, we predict a feature map. Next, the feature map is used in a clustering process. In the clustering results, each cluster can be considered as a detected part in that time step. After the clustering, all detected parts are incrementally matched between subsequent clustering results. Finally, the matching results in a set of part trajectories. As all components are incremental, new incoming observations can easily be attached to the current trajectory. These trajectories are then used in a downstream articulation mechanism estimator, which in our case is a factor graph (see \cref{subsec:methdology:factor_graph}) or \sturmauthor. 

\subsection{Motion Feature Prediction}
\label{subsubsec:methodology:category_independent_detection:feature_predictor}
Inspired by \cite{shao_motion-based_2018}, the motion feature predictor takes as input a pair of 6-channel RGB-XYZ images at time steps $t$ and $t+1$. The XYZ channels represent the $xyz$-coordinates of pixels in the camera frame, computed from the depth image. A ResNet-18 \cite{he_deep_2016} encoder takes the pair of images and outputs a low-dimensional feature map for each. The feature maps are then fused with upconvolutions with skip connections to generate a single dense motion map. This map has ten channels and holds a prediction of the following quantities for each of $n \in \{1, \dots, \text{width} \times \text{height} \}$ pixels:

% predict first component of our visual perception module takes as input four images, a pair of RGB $\mathcal{I} \in \realspace{\text{width} \times \text{height} \times 3}$ and XYZ $\mathcal{P} \in \realspace{\text{width} \times \text{height} \times 3}$ images representing camera frame coordinates at the current and the next time step. The images are fed into a shared (RGB/XYZ respectively) ResNet \cite{} encoder that extracts a spatially low-dimensional feature map for each image. We then fuse all feature maps through up convolutions with skip connections to generate our final motion feature map $\bm{m\!f}$. This map has 10 channels and holds a per pixel level prediction of the following quantities:
\begin{outline}
    \1 Importance $\beta_n^{(t)} \in \left[ 0, 1 \right]$ defined by the negative exponential distance to the image projected part center.
	\1 Geometric part center $\bm{c_n}^{(t)} \in \mathbb{R}^3$ in the camera frame.
	\1 Delta pose $\bm{\delta_n}^{(t)} \in se(3)$ of the part center from $t$ to $t+1$, represented in the camera frame.
\end{outline}
% Compared to our work, \cite{shao_motion-based_2018} additionally predicts a background segmentation mask and pixel-wise flow. 
While \cite{shao_motion-based_2018} predicts a translation in camera frame, the rotation is predicted in an object-centric frame. Compared to that, we predict the rotation also in the camera frame, allowing us to use the full delta pose for predicting the part center at the next time step $\bm{\tilde{c}_n}^{(t+1)} = \operatorname{Exp}(\bm{\delta_n}^{(t)})\, \bm{c_n}^{(t)}$.

% Talk about loss functions
The training loss is computed for each pixel as the sum:
\begin{align}
    \mathcal{L}
        &= \mathcal{L}_{\beta} + \beta(\gamma_c \mathcal{L}_{c} + \gamma_{\delta_v} \mathcal{L}_{\delta_v} + \gamma_{\delta_\omega} \mathcal{L}_{\delta_\omega} + \gamma_{\tilde{c}} \mathcal{L}_{\tilde{c}} \nonumber\\
        &\hspace{47pt}+ \gamma_{\sigma_c} \mathcal{L}_{\sigma_c} + \gamma_{\sigma_\delta} \mathcal{L}_{\sigma_\delta}).
\end{align}
$\mathcal{L}_{\beta}$, $\mathcal{L}_{c}$, $\mathcal{L}_{\delta_v}$, $\mathcal{L}_{\delta_\omega}$, and $\mathcal{L}_{\tilde{c}}$ are squared L2 losses on the error between the motion feature predictions and their ground truth values ($\bm{\delta_v}$ and $\bm{\delta_\omega}$ are the linear and angular components of $\bm{\delta}$, respectively). $\mathcal{L}_{\sigma_c}$ and $\mathcal{L}_{\sigma_\delta}$ are unsupervised losses on the squared L2 norm of the variances of $\bm{c}$ and $\bm{\delta}$, respectively. For each pixel, the variance is computed with neighboring features within a $3 \times 3$ window. These terms are designed to ensure the feature predictions are spatially consistent.

The $\gamma$ parameters can be tuned to scale the loss terms; for our experiments, we set $\gamma_{\delta_\omega} = 10$ and the rest to $1$. All the terms except $\mathcal{L}_\beta$ are scaled by the importance parameter $\beta$ to prevent unimportant pixels from influencing the loss.

\subsection{Motion Feature Clustering}
Because object parts are rigid, pixels belonging to the same part should output similar centers and delta poses. This step therefore clusters together pixels in the feature map that belong to the same part. First, we filter out unimportant pixels by selecting the $N$ pixels with the highest importance $\beta_n^{(t)}$. On this subset, we perform spectral clustering, inspired by \cite{huang_multibodysync_2021}. We construct an affinity matrix $\bm{A}^{(t)} \in \realspace{N \times N}$
\begin{align}
    A_{ij}^{(t)} = & \exp \frac{\|\bm{c_i}^{(t)} - \bm{c_j}^{(t)} \|^2}{-2 \sigma_A^2}
    + \exp \frac{\|\bm{\tilde{c}_i}^{(t+1)} - \bm{\tilde{c}_j}^{(t+1)} \|^2}{-2 \sigma_A^2},
\end{align}
where $i,j \in N$. $\sigma_A$ is a hyperparameter that controls how close centers should be to be considered part of the same cluster. We set $\sigma_A = 0.05\si{m}$.

We compute the singular value decomposition $\bm{U} \bm{\Sigma} \bm{V}^T = \bm{A}$ with singular values $\sigma_1, \dots \sigma_N$ in descending order. The number of ``significant" singular values are used to determine the number of clusters (i.e. parts) $K$. In other words, we want to find $K$ such that for all $i < K$, $\sigma_i \gg \sigma_K$. We determine $K$ by counting the singular values that are bigger than a fraction $\alpha$ of the sum of the first $M$ singular values:
\begin{align}
    K &= \left| \left\{ \sigma_i \mid \sigma_i \in \{\sigma_1, \dots, \sigma_N \},\, \sigma_i > \alpha \sum\nolimits_{m=1}^M \sigma_m \right\} \right|
\end{align}
$M$ is a hyperparameter which determines the maximum number of parts an object can have; we set $M=9$ as a reasonably high value, considering the maximum number of parts in our experiments is actually $2$. While \cite{huang_multibodysync_2021} uses a fixed $\alpha$ tuned specifically for the problem, we find this threshold dynamically at test time by building a histogram of $K$ computed over $100$ samples of $\alpha$ and choosing the most frequently occurring value of $K$.

Given $K$, we then perform $k$-means clustering on $\bm{U}$ with $K$ clusters, resulting in a set of pixels $N_k$ for each cluster $k \in K$. As each cluster represents a part, we compute the center and transformation for each part $k$ as an importance-weighted average over the pixels belonging to its cluster:
% First, we discretize the range $[0.0, 0.1]$ into $100$ possible values of $\alpha$. Next, for each threshold, we count the number of singular values that surpass it:
% \begin{equation}
% 	d_j = \left\vert \sigma_i > \alpha_l \varOmega \right\vert,~ j = 1 \dots 100.
% \end{equation}
% Since for each threshold we will get a value between 1 and 9, we can count the amount of occurrences in each bin, essentially giving us a discrete distribution over the amount of clusters. Eventually, we will pick the bin with the highest value, becoming our total parts detected $J$.
%Next, we will perform $K$-means clustering \cite{...} on $\bm{U}$ with $K=J$ being our detected part amount, giving us a cluster assignment $j$ for each of our $N$ pixels. Last, we will perform a weighted average for the center and transformation describing the motion of each part $j$
\begin{equation}
    \bm{c_k} = \dfrac{\sum_{n \in N_k} \beta_n \bm{c_n}}{\sum_{n \in N_k} \beta_n},
    \qquad \bm{\delta_k} = \dfrac{\sum_{n \in N_k} \beta_n \bm{\delta_n}}{\sum_{n \in N_k} \beta_n},
\end{equation}
where $N_k$ is the subset of $N$ pixels assigned to the $k$-th part.

\subsection{Cluster Matching}
The clusters are predicted from a pair of consecutive time steps, so they may not be temporally consistent over multiple time steps. The last step is to match clusters across time steps to turn motion features into a connected trajectory. We define a trajectory for part $k$ as a tuple $\xi_k = (C_k, \Delta_k)$, where $C = [\bm{c}^{(1)}, \dots, \bm{c}^{(T)}]$ is the sequence of part centers and $\Delta = [\bm{\delta}^{(1)}, \dots, \bm{\delta}^{(T)}]$ is the sequence of delta poses. The number of detected parts $K$ may change with each time step, so we keep a running list $\Xi = [\xi_1, \dots, \xi_L]$ of trajectories for all the $L$ parts ever detected.

% To that end, we first define a part trajectory as $\mathcal{Y}_j$ as an alternating, mixed sequence of centers and transformations
% \begin{equation*}
% 		\mathcal{Y}_j = \left[c_j^{(1)}, T_j^{(1)}, c_j^{(2)}, T_j^{(2)}, \ldots, c_j^{(T-1)}, T_j^{(T-1)},  c_j^{(T)} \right]
% \end{equation*}
% Throughout the whole matching process, we keep a set of ordered trajectories, initially filled with the detection results of the first time step pair.
% \begin{equation}
% 	    \mathbb{Y} = \left\{ \mathcal{Y}_j = \left[ c_j^{(1)}, T_j^{(1)} \right] \forall j \in J^{(1)} \right\}
% \end{equation}
For each incoming detection result $(\bm{c_k}^{(t)}, \bm{\delta_k}^{(t)})$ at time step $t$, we match the detection results to a trajectory $\xi_l \in \Xi$ by finding the one whose predicted center $\bm{\tilde{c}_l}^{(t)} = \operatorname{Exp}(\bm{\delta_l}^{(t-1)})~ \bm{c_l}^{(t-1)}$ is closest to the incoming one:
\begin{align}
    l_k
        &= \argmin\nolimits_{l=1}^L \left\| \bm{\tilde{c}_l}^{(t)} - \bm{c_k}^{(t)} \right\|.
\end{align}
If the number of detected parts $K$ is greater than the number of trajectories $L$ in $\Xi$, then for every unassigned part $k$, we assume it did not previously move and create a new trajectory $\xi_k = ([\bm{c_k}^{(t)}, \dots, \bm{c_k}^{(t)}], [\bm{0}, \dots, \bm{0}, \bm{\delta_k}^{(t)}])$ of length $t$. If a previously detected part $l$ is not detected at $t$, then we assume the part did not move and append $(\bm{c_l}^{(t-1)}, \bm{0})$ to $\xi_l$.
%attach an identity transformation and the center from the previous time step $t-1$. If we detect more parts than before, we assume the part did not move previously and add a new trajectory with length up to time $t$ with identity transformations and the incoming part center $c_{\hat{j}}^{(t)}$ to the end of our trajectory set.

For selecting the part trajectories in $\Xi$ to use in downstream articulation model estimation, we assume that the articulated body has a single joint with one fixed and one moving part. The fixed base part is assigned to the trajectory with the lowest variance, and the moving part is assigned to the one with the fewest missing detections across all the time steps. This logic can be easily extended to longer kinematic chains by selecting more trajectories in order of detection reliability.
% To select the base, we assign the trajectory with the lowest variance. As of now, we only consider objects with one static base part and one moving part, so we assign the object with the lowest variance as the base part and pick the first trajectory from the remaining trajectories (i.e. the one that was detected earliest) as our moving part.
%We can directly use these two trajectories as input to our proposed factor graph using the corresponding observation factors.

%\todo{1 page: ends at page 3}

%% file: tex/5_fg.tex
After detecting and tracking trajectories of part poses (e.g., with our visual perception module), the next step is to estimate the articulation model from the predicted part poses. We model this task as maximum a posteriori (MAP) estimation on a factor graph, an approach that has been shown to be extremely flexible for a range of estimation tasks in robotics \cite{sodhi_learningtactile_2021, yi_differentiable_2021}.
% For this task we will use factor graphs as they have shown great performance on a wide variery of robotic tasks
% compared to traditional filters 
% \todo{Cite 2-3 works @Brent? :)}.

\subsection{Factor Graph Background}
A factor graph is a bipartite graph that factorizes a probability distribution into conditionally-independent densities.
We define each factor graph using a set of latent variables $\bm{X}$, a set of observed variables $\bm{O}$ and a set of factors $\bm{\phi}$. All factors $i$ follow the same structure of 
\begin{equation}
    \phi_i \left(\bm{X}_i ; \bm{O}_i\right) = \exp \left( - \frac{1}{2} \| \bm{r}_i \left(\bm{X}_i ; \bm{O}_i \right)\|_{\Sigma_i}  \right)
\end{equation}
where $\bm{X}_i \subseteq \bm{X}$ and $\bm{O}_i \subseteq \bm{O}$ are defined by the connectivity of the graph. $\| \bm{r}_i \left(\bm{X}_i ; \bm{O}_i \right)\|_{\Sigma_i}$ is the Mahalanobis distance $d_i$ with covariance matrix $\Sigma_i$ of a residual function $\bm{r}_i (\bullet)$. If a factor $\phi_i$ does not use any observations, $\bm{O}_i$ will be the empty set $\emptyset$ and for clarity we will drop $\bm{O}_i$ from the parameter list.

\begin{figure}[t]
    \centering
    \begin{subfigure}[t]{0.49\linewidth}
        \centering
        \includegraphics[width=\textwidth]{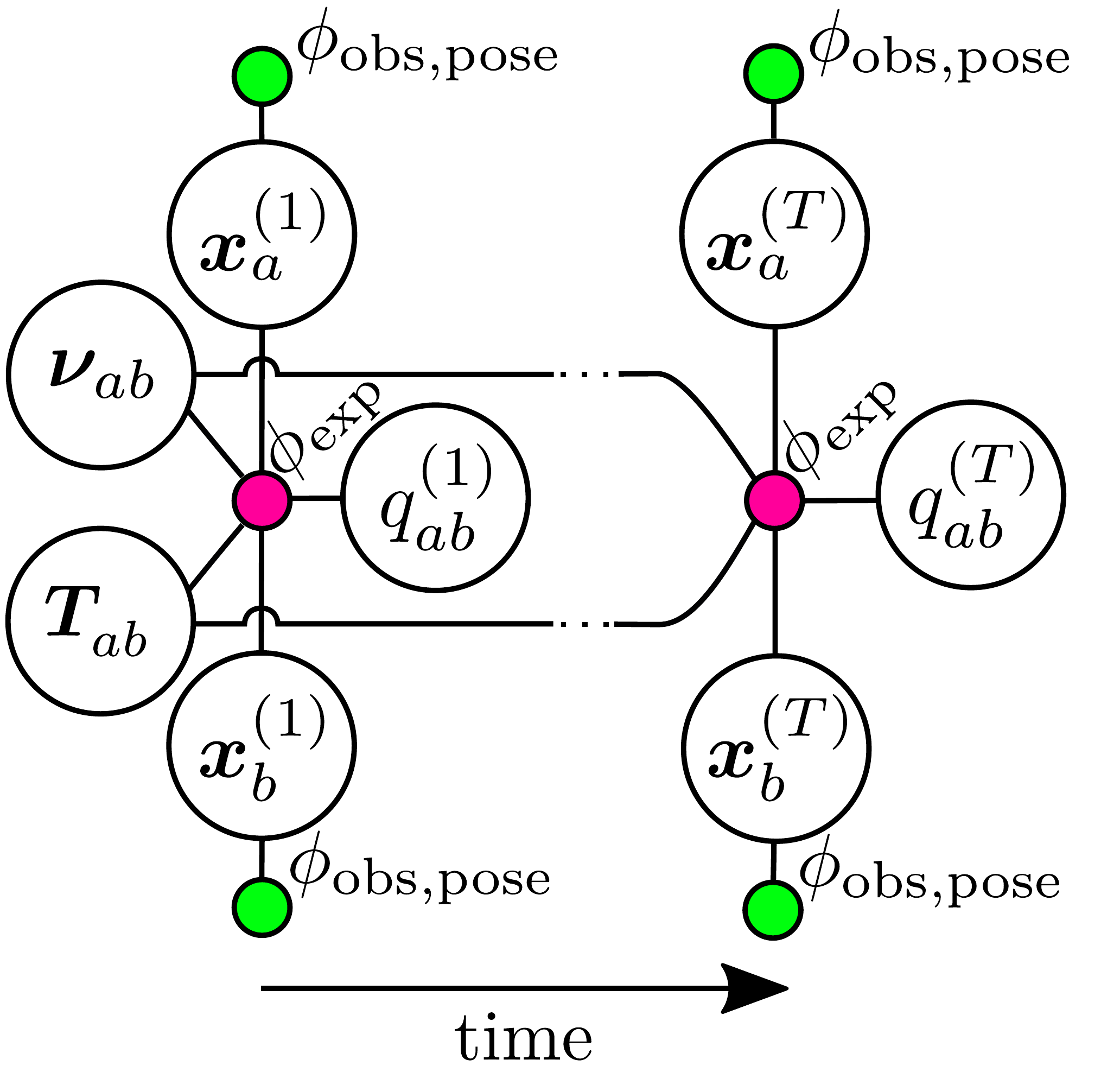}
        \caption{\footnotesize Pose observations}
        \label{subfig:factor_graph:poses}
    \end{subfigure}%
    \hfill
    \begin{subfigure}[t]{0.49\linewidth}
        \centering
        \includegraphics[width=\textwidth]{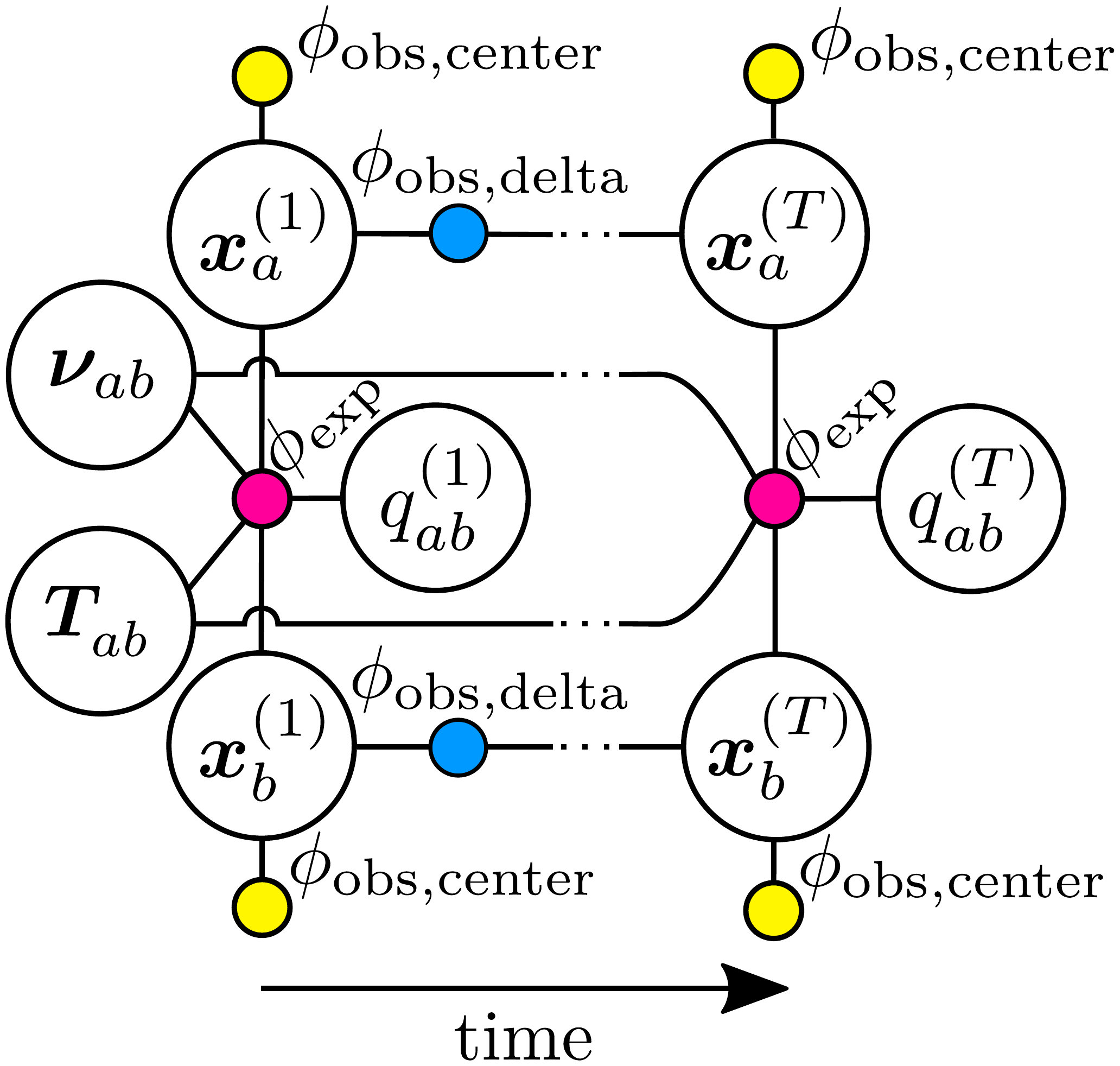}
        \caption{\footnotesize Center/delta pose observations}
        \label{subfig:factor_graph:transformations}
    \end{subfigure}
    \caption{Our two factor graphs with different observation inputs, indicated by $\phi_{obs,\bullet}$ factors. Both factor graphs infer the same latent variables (white nodes). The exponential factor $\phi_{exp}$ connects the time-varying latent variables $\bm{x}^{(t)}$ and $q^{(t)}$ to the joint parameters $\bm{\nu}$ and $\bm{T_{ab}}$.}
    \label{fig:factor_graph}
\end{figure}

% Abstractly speaking, a joint describes the motion between the poses of two rigid parts $\poselat[][i]$ and $\poselat[][j]$ in space constrained by its configuration space. A joint is then parameterized by its time-varying joint states $\bm{q}_{ij}^{(t)}$ and its time-constant joint parameters $\bm{\Theta}_{ij}$, for which their respective spaces depend on the underlying joint model $\mathcal{M}_{ij}$. A general joint model $\mathcal{M}$ then can be expressed as a function of its joint state and parameters
% \begin{equation*}
%     f_{\mathcal{M}}( \bm{q}^{(t)} ; \bm{\Theta} )
% \end{equation*}
% that returns a transformation in $SE(3)$. 

% For example, for a prismatic joint, the joint state $q$ is one dimensional $\realspace{1}$ and the required joint parameters $\bm{\Theta} = \left\{ \transform[][\text{base}], \bm{v} \right\}$ are a set of a rigid base transformation $\transform[][\text{base}] \in SE(3)$ and a translation vector $\bm{v} \in \realspace{3}$\cite{sturm_probabilistic_2011}. For a prismatic joint $\mathcal{M} = \text{prismatic}$ the joint function is then defined through
% \begin{equation*}
%     f_{\text{prismatic}} \left(q; \transform[][\text{base}], \bm{v} \right) = \transform[][\text{base}] \text{trans} \left( q \bm{v} \right)
% \end{equation*}
% where $\text{trans}\left( \bullet \right)$ is a function that constructs a transformation matrix given a translation \cite{sturm_probabilistic_2011}.

To improve convergence characteristics, we perform MAP estimation over the distribution specified by the factor graph using the Levenberg-Marquardt algorithm \cite[Sec.4.7.3]{Gill81} on $P$ parallel, randomly initialized instances of the problem, and then select the solution with the lowest cost. For all of our experiments, we use $P=10$. For robustness against outliers, we use a Huber loss wrapped around the Mahalanobis distance $d_i$ of each factor with a small $\delta=0.01$:

\begin{equation}
    L_\delta (d_i) = \begin{cases}
     \frac{1}{2}{d_i^2}                   & \text{for } |d_i| \le \delta, \\
     \delta (|d_i| - \frac{1}{2}\delta), & \text{otherwise.}
    \end{cases}
\end{equation}

\subsection{Proposed Factor Graph}

% \begin{figure*}
%     \centering
%     \includegraphics[width=.7\textwidth]{assets/img/full_graph.png}
% \end{figure*}
In Fig.~\ref{fig:factor_graph}, we propose two factor graphs, which share the same latent structure but have different observations attached. We will now first explain the shared latent structure and then how the different observations are incorporated. 

% \subsubsection{Latent Structure}
Both factor graphs contain latent part poses $\poselat[][]_k^{(t)}$, one pose per detected part $k \in \left[1, K\right]$ at each time step $t \in \left[1, T\right] $. For each pair of parts $(a, b)$ that are connected through a joint, we aim to estimate a sequence of time-varying joint configurations $q_{ab}^{(t)}$ and a pair of time-invariant joint parameters $( \bm{\nu_{ab}}, \bm{T_{ab}})$.

Similarly to \cite{jain_screwnet_2021}, we parameterize the joint with a twist $\bm{\nu} = (\bm{v}, \bm{\omega})$ that is able to express rigid, prismatic, revolute and helical joint types. However, while \cite{jain_screwnet_2021} uses two scalar parameters to scale $\bm{v}$ and $\bm{\omega}$ separately, we use a single joint configuration parameter $q^{(t)} \in \mathbb{R}$ to scale the entire twist, as described in Eq.~\ref{eqn:twist_joint_formulation}.

To allow arbitrary local frame placements for the articulated parts, we incorporate an additional transformation $\transform[][\text{twist}] \in SE(3)$ from the twist frame to the camera frame, as done in \cite{sturm_probabilistic_2011}. The joint articulation function then becomes:
%Similar to \cite{jain_screwnet_2021} we derive a joint model $\mathcal{M}=\text{twist}$ that uses a twist formulation to represent the aforementioned joint types through the same mathematical formulation. Following \cite{lynch_modern_2017} we directly express the twist by $\bm{\nu} = \left( \bm{v}, \bm{\omega} \right) \in se(3)$ where $\bm{v}$ corresponds to the translation and $\bm{\omega}$ to the rotation component. Compared to \cite{jain_screwnet_2021}, we will use a single joint state $q^{(t)}$ to scale the twist and not two separate ones to express the translation and rotation in isolation. Additionally, to allow arbitrary frame placements for the parts we need to incorporate an additional base transformation $\transform[][\text{base}]$ as done in \cite{sturm_probabilistic_2011}.
\begin{align}
    f_{\text{twist}}\left( q^{(t)} ; \bm{\nu}, \transform[][\text{twist}] \right) = \transform[][\text{twist}] \lieExp{
		 q^{(t)} \bm{\nu}}.
\end{align}

At each time step, the joint variables and latent poses are connected through the exponential factor $\phi_{exp}$. This factor compares the relative transformation between the latent part poses $\left(\poselat[][a] \right)^{-1} \poselat[][b]$ to the expected transformation computed by the joint function $f_{\text{twist}}(q_{ab}; \bm{\nu_{ab}}, \bm{T_{ab}})$. The residual error is computed by:
\begin{align}
    % &\bm{r}_{\text{exp}}(\bm{x_a}, \bm{x_b}, \bm{T_{ab}}, \bm{\nu_{ab}}, q_{ab}) \nonumber \\
    &\bm{r}_{\text{exp}} \left(\bm{X}_{\text{exp}} = \left\{ \bm{x_a}, \bm{x_b}, \bm{T_{ab}}, \bm{\nu_{ab}}, q_{ab} \right\} \right) \nonumber \\
    &\quad= f_{\text{twist}}(q_{ab}; \bm{\nu_{ab}}, \bm{T_{ab}}) \ominus (\bm{x_a}^{-1} \bm{x_b}),
\end{align}
where $\ominus$ is the twist error between two poses $\bm{x}, \bm{y} \in SE(3)$:
\begin{align}
    \bm{x} \ominus \bm{y}
        &= \operatorname{Log}(\bm{x}^{-1} \bm{y}) \in se(3).
\end{align}
% \begin{align}
% 	&\bm{r}_{\text{exp}}(
% 		\transform[i][\text{tw},j], \bm{\nu}, q_{ij}^{(t)}, \poselat[][i], \poselat[][j]
% 	) \nonumber \\
% 	&\quad= f_{\text{twist}}( q_{ij}^{(t)} ; \transform[i][\text{tw},j], \bm{\nu}_{ij} ) \ominus \left( \left(\poselat[][i] \right)^{-1} \poselat[][j] \right)
% 	\nonumber \\
% 	\label{eqn:exp_factor_resid}
% 	&\quad= \left(\transform[i][\text{tw},j] \lieExp{
% 		\bm{\nu}_{ij} q_{ij}^{(t)}}
% 	\right) \ominus \left( \left(\poselat[][i] \right)^{-1} \poselat[][j] \right)
% \end{align}
% Hence, the set of all latent variables inferred by our proposed factor graph is given as $\bm{X} = \{ \bm{x_a}, \bm{x_b}, q_{ab} \forall t \in T \and \bm{T_{ab}}, \bm{\nu_{ab}}\}$

%\subsection{Observation Factors}
The factor graph can infer its latent variables from different observations, depending on what the upstream perception method predicts. Thus, the factor graph could be used in conjunction with a wide range of perception methods. In the following, we highlight three possible observation types: pose observations (e.g., from CAPTRA \cite{weng_captra_2021}), centers, and pose changes (e.g., from our proposed part tracking method).
%Depending on our perception method our factor graph allows us to use different observation modalities. In the following we will highlight three possible modalities.
\subsubsection{Part Pose Observation}
If the perception method tracks full 6D poses for each articulated body part, we can use the observation factor $\phi_{\text{obs}, \text{pose}}$ to compare an observed pose $\poseobs[][] \in SE(3)$ to a latent pose $\poselat[][]$ with the residual
\begin{align}
	&\bm{r}_{\text{obs}, \text{pose}}
	\left(
% 		\poselat[][]; \poseobs[][]
        \bm{X}_{\text{obs},\text{pose}} = \left\{\poselat[][] \right\} ;  \bm{O}_{\text{obs},\text{pose}} = \left\{ \poseobs[][] \right\}
	\right) \nonumber \\
	&\quad = \poselat[][] \ominus \poseobs[][].
\end{align}

\subsubsection{Part Center Observation}
If the perception method only tracks the positions of part centers, we use the part center observation factor $\phi_{\text{obs}, \text{center}}$ to compare the observed part center $\bm{c}$ to the position portion of the latent part pose $\poselat[][]_p$ with the residual
\begin{align}
	&\bm{r}_{\text{obs}, \text{center}}
	\left(
	   % \poselat[][]; \bm{c}
		\bm{X}_{\text{obs}, \text{pose}} = \left\{\poselat[][] \right\}; \bm{O}_{\text{obs},\text{center}} = \left\{\bm{c} \right\}
	\right) \nonumber \\
	&\quad= \poselat[][]_p  - \bm{c}.
\end{align}

\subsubsection{Part Pose Change Observation}
If the perception method tracks the changes in part poses between consecutive time steps, we use the delta pose observation factor $\phi_{\text{obs}, \text{delta}}$ to compare the observed delta pose $\bm{\delta}^{(t)} \in se(3)$ expressed in the world frame with the latent poses $\poselat[][]^{(t)}$ and $\poselat[][]^{(t+1)}$:
\begin{align}
    &\bm{r}_{\text{obs}, \text{delta}}
	\left(
% 		\poselat[][]^{(t)}, \poselat[][]^{(t+1)}; \bm{\delta}^{(t)}
        \bm{X}_{\text{obs}, \text{delta}} = \left\{\poselat[][]^{(t)}, \poselat[][]^{(t+1)} \right\} ; \bm{O}_{\text{obs},\text{delta}} = \left\{ \bm{\delta}^{(t)} \right\}
	\right) \nonumber \\
    &\quad = \left( \operatorname{Exp}(\bm{\delta}^{(t)})~ \poselat[][]^{(t)} \right) \ominus \poselat[][]^{(t+1)}.
\end{align}
%The last observation factor we introduce, $\phi_{\text{obs}, \text{trans}}$, compares the relative transformation between two subsequent latent poses $\poselat[][]^{(t)}$ and $\poselat[][]^{(t+1)}$. The relative transformation $\bm{Z}^{(t)}$ is observed as a twist in the world frame we are acting in. Therefore, we pre-multiply the exponentiated version. The resulting residual vector has the form
% \begin{align*}
%     &\bm{r}_{\text{obs}, \text{center}}
% 	\left(
% 		\poselat[][]^{(t)}, \poselat[][]^{(t+1)}; \transformobs[][]^{(t)}
% 	\right)\\
% 	&\quad=	\left( \lieExp{\transformobs[][]^{(t)}} \poselat[][]^{(t)} \right) \ominus \poselat[][]^{(t+1)}
% \end{align*}
% An equivalent formulation that first transforms the observed twist to the local part frame is
% \begin{align*}
%     &\bm{r}_{\text{obs}, \text{center}}
% 	\left(
% 		\poselat[][]^{(t)}, \poselat[][]^{(t+1)}; \transformobs[][]^{(t)}
% 	\right)\\
% 	&\quad= \left( \poselat[][]^{(t)} \lieExp{\vec{Ad}_{ \left( \poselat[][]^{(t)} \right)^{-1}}
% 	    \left( 
% 	        \transformobs[][]^{(t)}
% 	    \right)} \right)
% 	    \ominus \poselat[][]^{(t+1)}
% \end{align*}

%\todo{1 page: ends at page 4}

%% file: tex/6_metric.tex
\begin{figure}
    \centering
    \includegraphics[width=\columnwidth]{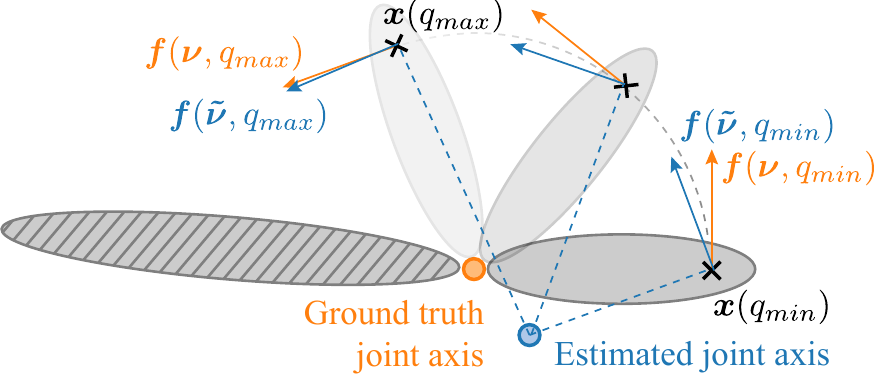}
    \caption{The tangent similarity metric measures the cosine similarity between the predicted linear velocity $\bm{f}(\bm{\tilde{\nu}}, q)$ and ground truth $\bm{f}(\bm{\nu}, q)$ along the path traced by the grasping point $\bm{x}(q)$.}
    \label{fig:metric}
\end{figure}

We introduce a new error metric for articulation models that captures how useful a prediction is for robot manipulation. A commonly used metric is the angle error of the predicted joint axis. However, when the joint type is unknown, this metric can be misleading; the predicted joint axis could have 0 angle error, but if the predicted joint type (i.e. prismatic vs. revolute) is wrong, the predicted axis actually captures an orthogonal range of motion and thus is useless for manipulation. Meanwhile, a prediction with the incorrect joint type and axis may actually be sufficient to manipulate the object if the tangent direction points in the correct direction. Our ultimate goal is to manipulate articulated objects, so we propose a metric that measures how well a robot would be able to manipulate the articulated object given an estimated articulation model.

More formally, suppose we have two rigid bodies connected by a single twist joint parameterized by $\vec{\nu} = (\vec{v}, \vec{\omega}) \in se(3)$. We assume one body is a rigid base, and we want to manipulate the second body by grasping it at a given fixed point and pulling it such that the joint configuration $q$ goes from $q_{min}$ to $q_{max}$. Let $\vec{x_0}$ be the grasping point when $q = 0$. The grasping point follows the path
\begin{align}
    \vec{x}(q)
        &= \lieExp{q \vec{\nu}} \vec{x_0} \in \mathbb{R}^3 \label{eq:path}
\end{align}
for the range $q \in [q_{min}, q_{max}]$. A visualization is presented in Fig. \ref{fig:metric}.

At test time, the joint motion is unknown, so we need to predict it. Let $\vec{\tilde{\nu}} = (\vec{\tilde{v}}, \vec{\tilde{\omega}}) \in se(3)$ be a prediction of the joint motion. We assume that we want to manipulate the articulated body using the predicted joint motion with a rigid grasp at the grasping point.

We define the tangent similarity metric as the average cosine similarity between the predicted and true linear velocities $\vec{\tilde{v}}$ and $\vec{v}$ at the grasping point along the path $\vec{x}(q)$:
\begin{align*}
    J(\vec{\nu}, \vec{\tilde{\nu}})
        &= \frac{1}{q_{max} - q_{min}} \int_{q_{min}}^{q_{max}} \frac{\vec{f}(\vec{\nu}, q)}{\|\vec{f}(\vec{\nu}, q)\|} \cdot \frac{\vec{f}(\vec{\tilde{\nu}}, q)}{\|\vec{f}(\vec{\tilde{\nu}}, q)\|} ~dq 
\end{align*} 
where
\begin{align}
    \vec{f}(\vec{\nu}, q)
        &= \vec{Ad}_{\lieExp{\vec{x}(q)}^{-1}}(\vec{\nu})_{\vec{v}} \label{eq:linear_motion} \\
        &= \vec{v} + \crossm{\vec{\omega}} \vec{x}(q) \nonumber.
\end{align}

The adjoint operator $\vec{Ad}_{\lieExp{\vec{x}(q)}^{-1}} : se(3) \rightarrow se(3)$ takes the twist $\vec{\nu}$ and transforms it to the local frame of the grasping point $\vec{x}(q) \in \mathbb{R}^{3}$ via the transformation $\lieExp{\vec{x}(q)}^{-1} \in SE(3)$. For this metric, we are only concerned with the linear component $\vec{\nu}_{\vec{v}}$ of the resulting twist. A perfect prediction yields $1$, while an orthogonal prediction yields $0$.

While the integral does not have a closed form solution, we can easily approximate it with equally spaced samples of $q$ in the range $[q_{min}, q_{max}]$; for the joint ranges used in our experiments, we found $100$ samples was enough to accurately estimate within $\pm 0.001$.

This metric is physically meaningful, since it measures how much of an applied force would be able to move the second body in the direction of joint motion. If a force that is constantly $60^\circ$ off is applied to the second body, $\cos(60^\circ) = 0.5$. Therefore, half of the applied force would still be in the direction of joint motion and this metric would yield $0.5$. In other words, this component of the applied force could move the second body with half the effectiveness.

% \subsubsection{Linear Motion Angle Error}

% This metric measures the average angle error between the predicted and true linear velocities $\vec{\tilde{\nu}}$ and $\vec{\nu}$ at the grasping point along the path $\vec{x}(q)$. This is more in line with existing metrics to measure joint axis prediction accuracy. Unlike existing metrics, it does not suffer from incorrect categorization of joint types. $0^\circ$ means the prediction is perfect and the robot will be able to manipulate the articulated body perfectly, while $90^\circ$ means the prediction is orthogonal and the robot will not be able to move the articulated body at all.
% \begin{align}
%     J(\vec{\nu}, \vec{\tilde{\nu}})
%         &= \frac{1}{q_{max} - q_{min}} \notag \\
%         &\enskip \int_{q_{min}}^{q_{max}} \arccos \left( \frac{\vec{f}(\vec{\nu}, q)}{\|\vec{f}(\vec{\nu}, q)\|} \cdot \frac{\vec{f}(\vec{\tilde{\nu}}, q)}{\|\vec{f}(\vec{\tilde{\nu}}, q)\|} \right) dq
% \end{align}

% $\vec{f}(\vec{\nu}, q)$ is the same as the one defined in Eq.~\ref{eq:linear_motion}.

%\todo{1 col: ends at page 5 col 1}

%% file: tex/7a_fg_exp.tex
\begin{figure*}[t!]
    \centering
    \includegraphics[width=\textwidth]{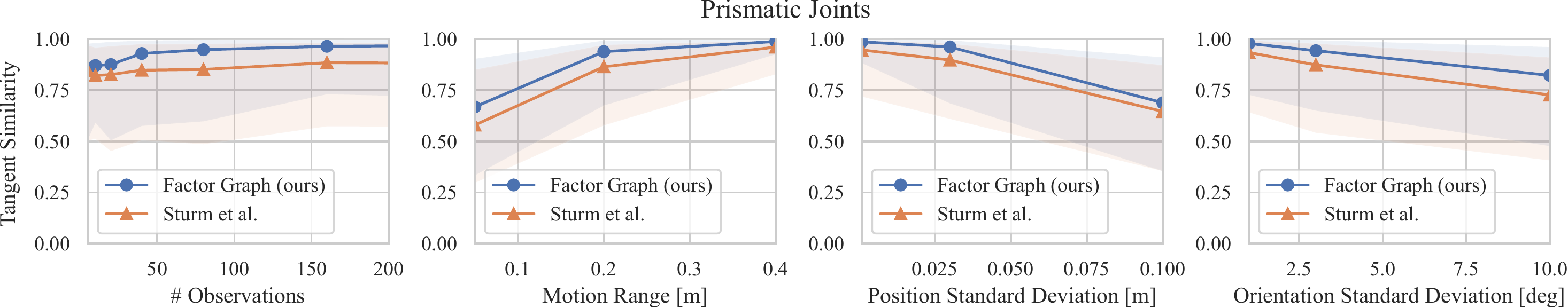}
    
    \vspace{0.2cm}
    
    \includegraphics[width=\textwidth]{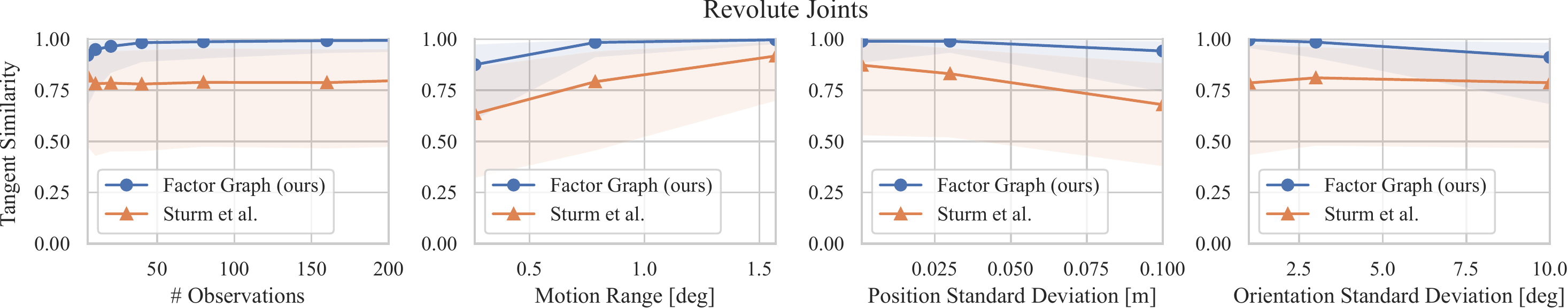}
    \caption{Comparison between articulation model estimation methods (\textbf{FG}, \textbf{Sturm}) on synthetically generated noisy data. The lines represent the median tangent similarity score (1.0 represents a perfect prediction) over 50 runs, while the shaded region represents the 25th and 75th percentiles. The first two columns show that both methods benefit from an increased number of observations and larger motion ranges, but $\textbf{FG}$ is able to achieve nearly perfect scores with fewer observations and smaller motion ranges. The last two columns show that $\textbf{FG}$ is more robust to position and orientation noise.}
    \label{fig:only_poses}
\end{figure*}

%Our articulation prediction pipeline is comprised of two modules: 1) the visual perception module that tracks part poses, and 2) the factor graph that predicts joint articulation from the part poses.
In the experiments, we test two hypotheses. H1: The factor graph is able to outperform baselines as a standalone articulation module estimation module when given noisy pose data. H2: Explicitly tracking part poses allows the full tracking and estimation pipeline to generalize across object categories.

\subsection{Factor Graph Experiment (H1)}

In this experiment, we evaluate the factor graph's ability to handle noisy pose observations. The factor graph (\textbf{FG}) can be used as a standalone module that estimates articulation models from part poses, just like \sturmauthor (\textbf{Sturm}). Therefore, we use \textbf{Sturm} as a baseline evaluating on two types of pose data: 1) synthetically generated data with controlled noise and 2) predicted poses from a state-of-the-art category-level part tracking method, CAPTRA~\cite{weng_captra_2021}. Since CAPTRA assumes that the joint type is already known, the role of \textbf{FG} and \textbf{Sturm} when used in conjunction with CAPTRA is to solely estimate the joint parameters for the known joint type. The purpose of this experiment is to show how the methods handle predictions from a learned system with non-Gaussian noise characteristics.

\subsubsection{Setup}

\begin{table}
    \centering
    \begin{tabular}{|c|c|c|c|}
        \hline
        Variable & Values \\
        \hhline{|=|=|}
        %Joint type & $\{\text{prismatic}, \text{revolute} \} $ \\
        %\hline
        \# observations $T$ & $\{ 5, 10, 20, 40, 80, 160, 320 \}$ \\
        \hline
        Motion range $q_{\max}$ & $
            \begin{array}{rl}
               \text{revolute:} & \{ 15 \degree, 45 \degree, 90 \degree \} \\
                \text{prismatic:} & \{ 0.05 \si{m}, 0.2 \si{m}, 0.4 \si{m} \} \\
            \end{array}
        $ \\
        \hline
        Position noise $\sigma_{\text{pos}}$ & $ \{ 0.001 \si{m}, 0.03 \si{m}, 0.1 \si{m} \}$ \\
        \hline
        Orientation noise $\sigma_{\text{ori}}$ & $\{ 1.0 \degree, 3.0 \degree, 10.0 \degree \}$ \\
        \hline
    \end{tabular}
    \caption{Parameters used for the synthetic pose experiment.}
    \label{tab:noisy_poses:parameter_overview}
\end{table}

\begin{figure}
    \centering
    \includegraphics[width=\columnwidth]{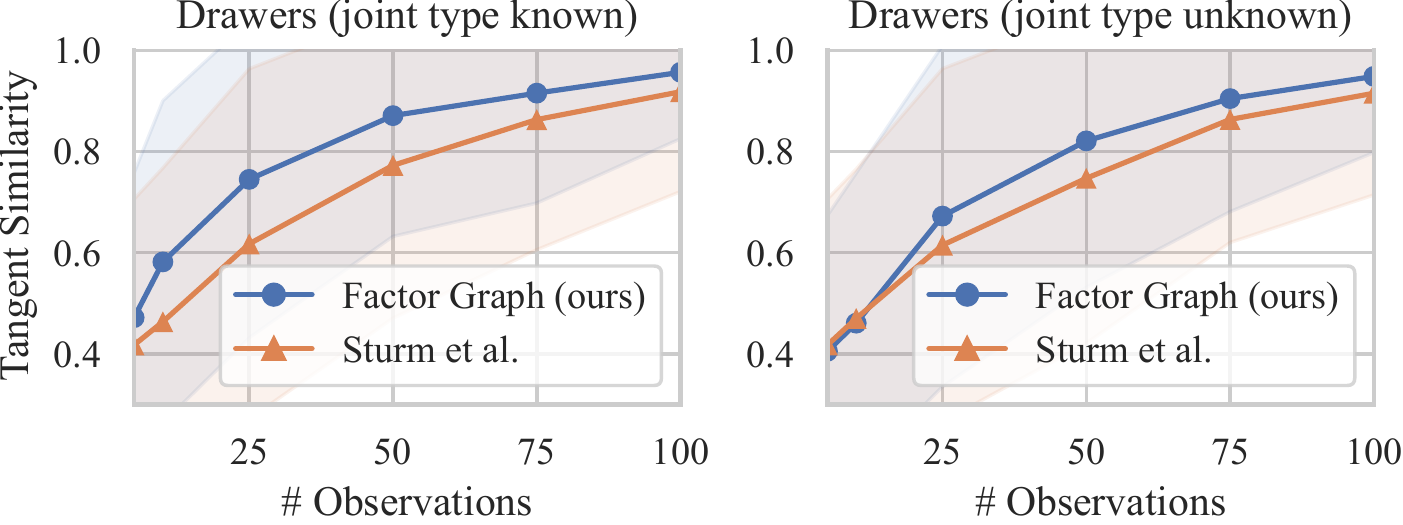}
    
    \vspace{0.2cm}
    
    \includegraphics[width=\columnwidth]{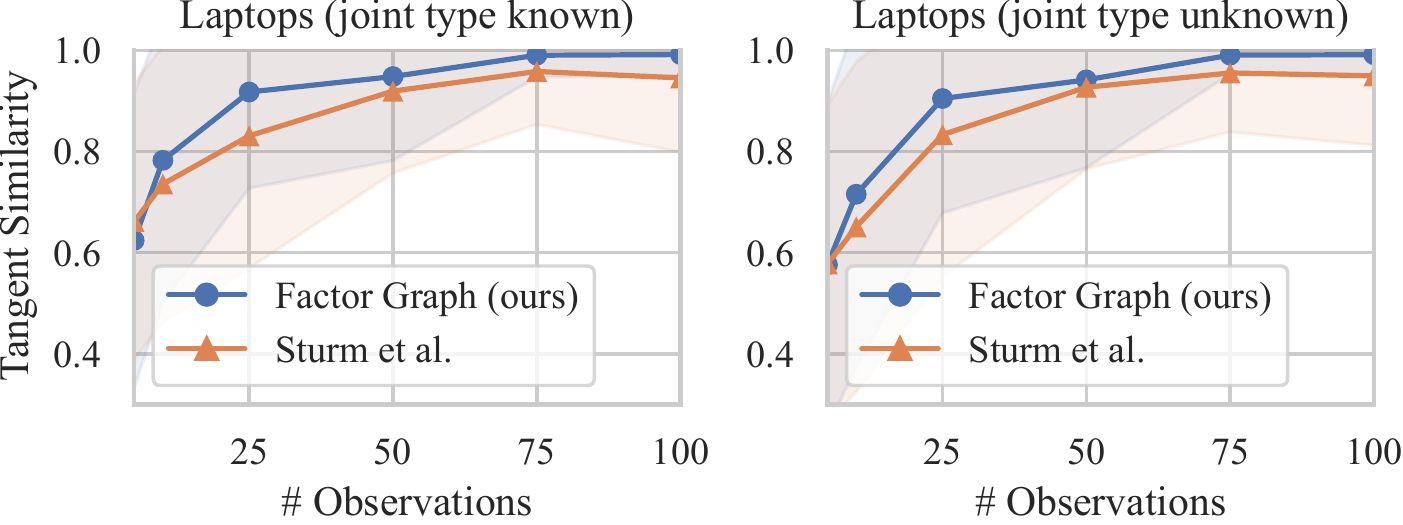}
    % \begin{subfigure}[t]{0.49\linewidth}
    %     \centering
    %     \includegraphics[width=\textwidth]{assets/img/results/CAPTRA/drawer_all_obs.png}
    %     \caption{...}
    %     \label{subfig:captra_results:drawers}
    % \end{subfigure}%
    % \hfill
    % \begin{subfigure}[t]{0.49\linewidth}
    %     \centering
    %     \includegraphics[width=\textwidth]{assets/img/results/CAPTRA/laptop_all_obs.png}
    %     \caption{...}
    %     \label{subfig:captra_results:laptop}
    % \end{subfigure}
    \caption{Articulation model estimation using part poses provided by CAPTRA. Since CAPTRA assumes the joint type is known, we test \textbf{FG} and \textbf{Sturm} with and without the known joint type constraints. We plot the mean and standard deviation tangent similarity scores over 60 simulation runs.}
    \label{fig:captra_results}
\end{figure}

For \textbf{Sturm}, we use the original implementation of \sturmauthor with its provided parameters. During initial testing, \textbf{Sturm} sometimes misclassified joints as rigid; since the joints are always either prismatic or revolute in our experiments, we removed rigid joints from \textbf{Sturm} for a fair comparison. \textbf{Sturm} takes $6$-DoF part poses as input, so we use $\textbf{FG}$ with pose observation factors (Fig.~\ref{subfig:factor_graph:poses}). We set the noise parameters for both \textbf{FG} and \textbf{Sturm} to the same noise parameters used to generate the synthetic and CAPTRA data.
%Additionally, we initialize the factor graph optimization by setting the initial poses to the observed poses.

To generate synthetic pose data, we vary four parameters: the number of observations, the range of the articulated motion, the position noise, and the orientation noise. The exact parameter values are given in \Cref{tab:noisy_poses:parameter_overview}.
\begin{appendixcond}
Details on how the noisy poses are generated are included in Appx.~\ref{appendix:fg_experiment}.
\end{appendixcond}

For CAPTRA data, we use 60 simulation runs each for two categories, \texttt{Drawer} and \texttt{Laptop}, from the simulated SAPIEN dataset \cite{weng_captra_2021} on which CAPTRA was tested.
%This dataset contains two \texttt{Drawer} and six \texttt{Laptop} instances, with 10 simulation runs per instance. Since the \texttt{Drawer} consists of three moving parts, we treat each moving part as a separate simulation run, giving us 60 total simulation runs per category.

% Additionally, we evaluated our method on a real world data where robot opens a drawer (Section F.4 \cite{weng_captra_2021}). 
%The full sequence of observations has a length of 100. To test sample efficiency, we sub-sample different amounts of observations from the full sequence, but always include the first and last one. We sub-sampled $5, 10, 25, 50, 75$ observations as well as tested on the full sequence with 100 observations.

Since CAPTRA assumes that the joint type is known, we also evaluate \textbf{FG} and \textbf{Sturm} with the joint type given. For \textbf{FG}, this means adding a constraint to ensure that the predicted twist $\bm{\nu} = (\bm{v}, \bm{\omega})$ obeys the given joint constraint: $\bm{\omega} = \bm{0}$ for prismatic joints and $\bm{v} \times \bm{\omega} = \bm{0}$ for revolute joints.

\subsubsection{Results}

The results for the synthetic pose experiment are shown in Fig.~\ref{fig:only_poses}. \textbf{FG} outperforms \textbf{Sturm} for both joint types, but especially for revolute joints. \textbf{Sturm} performs worse with revolute joints because it has a tendency to misclassify revolute joints as prismatic.

The results for the CAPTRA pose experiment are shown in Fig.~\ref{fig:captra_results}. \textbf{FG} performs better than \textbf{Sturm} for the \texttt{Laptop} category, and achieves high scores with fewer observations (i.e. better sample efficiency) for the \texttt{Drawer} category. \textbf{Sturm} did not benefit from knowing the joint type, while \textbf{FG} did; this information helped improve its sample efficiency for the \texttt{Drawer} category. Both methods performed worse for drawers than laptops, likely due to the fact that CAPTRA prediction errors were larger for \texttt{Drawer} parts ($0.465 \si{cm}$) than for \texttt{Laptop} parts ($0.335 \si{cm}$).

\begin{figure}
    \centering
    \begin{subfigure}{.325\columnwidth}
        \includegraphics[angle=270,width=\linewidth]{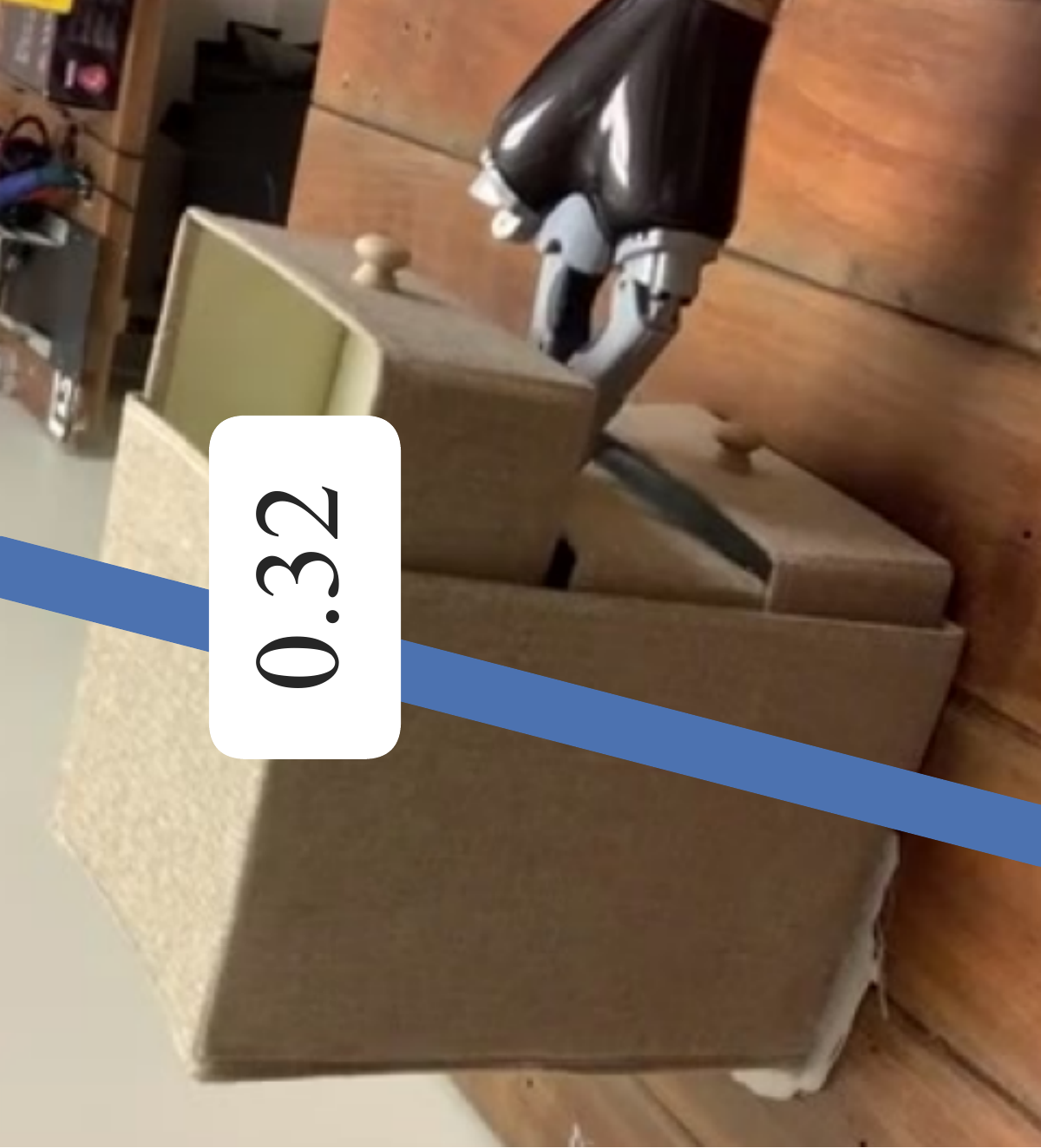}
        \caption{$t=0\si{s}$}
    \end{subfigure}
    \begin{subfigure}{.325\columnwidth}
        \includegraphics[angle=270,width=\linewidth]{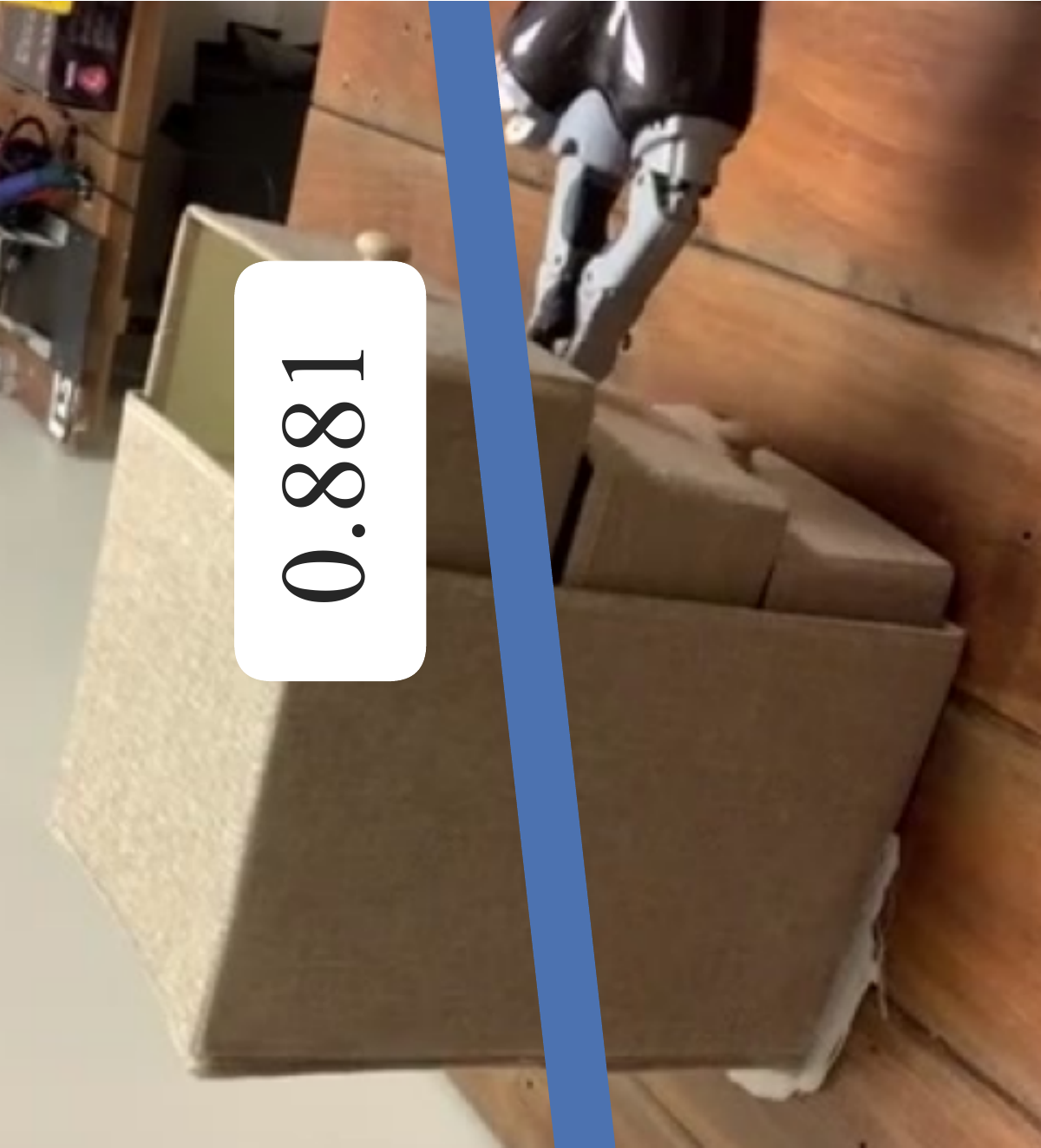}
        \caption{$t=1\si{s}$}
    \end{subfigure}
    \begin{subfigure}{.325\columnwidth}
        \includegraphics[angle=270,width=\linewidth]{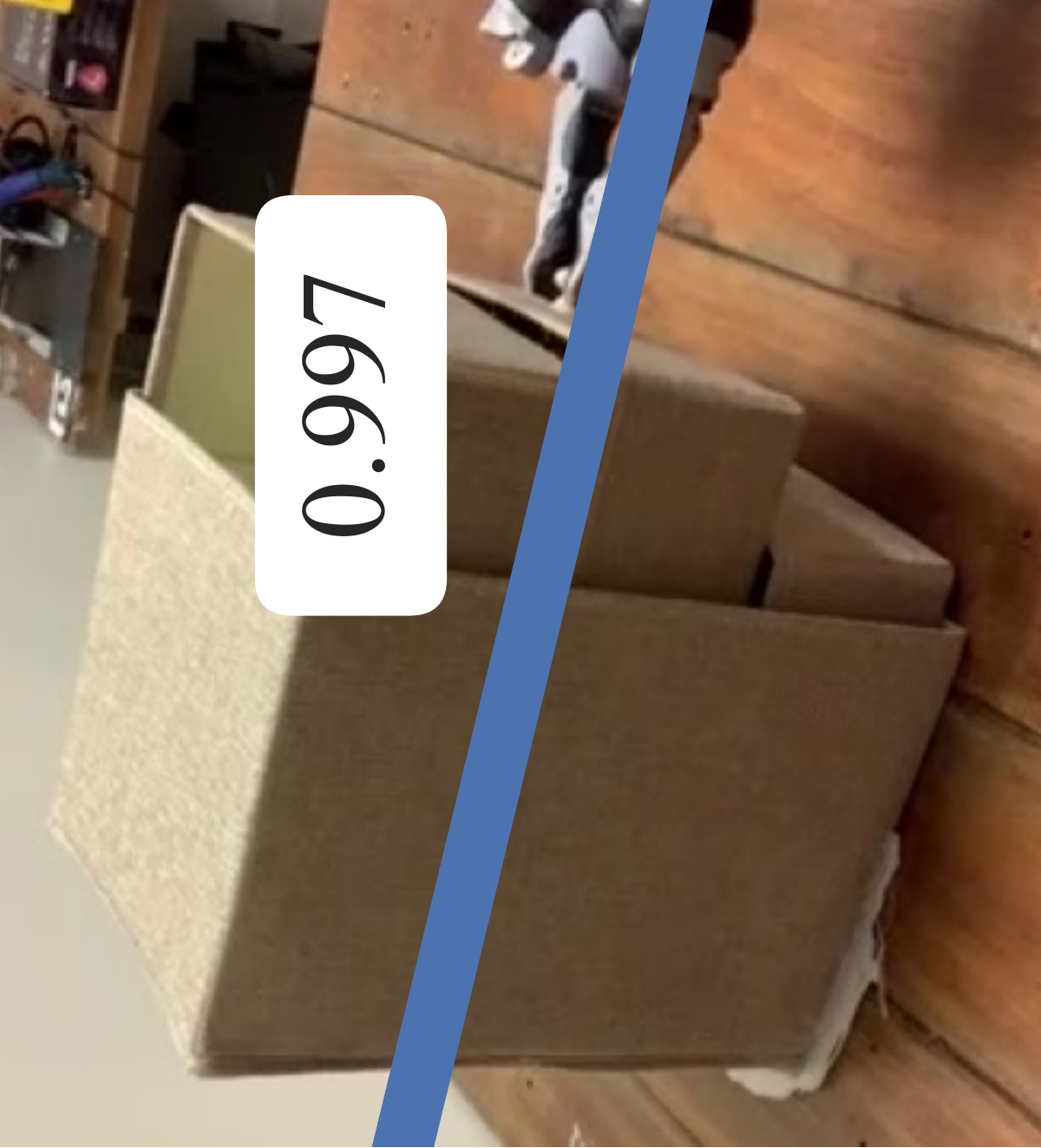}
        \caption{$t=2\si{s}$}
    \end{subfigure}
    \caption{Real world results using CAPTRA for part pose tracking and \textbf{FG} for articulation model estimation. The blue line shows the estimated joint axis. As the robot interacts with the object, the estimate becomes more accurate, as indicated by the tangent similarity score shown in the white box.}
    \label{fig:captra_real}
\end{figure}

Fig.~\ref{fig:captra_real} shows qualitative real world results using CAPTRA to track part poses of a drawer. %\footnote{CAPTRA tracking results provided by Yijia Weng.}

%\todo{1 page: ends at page 6 col 1}

%% file: tex/7b_full_exp.tex
\subsection{Full Pipeline Experiment (H2)}

\begin{table}[]
    \centering
    \begin{tabular}{c|c|c}
        \multicolumn{2}{c|}{Method} & \multirow{2}{6em}{\centering Prior Visual Information} \\
        \cline{1-2}
		Visual Perception & Estimation & \\
		\hhline{=|=|=}
        Ours (\textbf{PT}) & Factor Graph (\textbf{FG}) & None\\
        \hline
        Ours (\textbf{PT}) & \sturmauthor (\textbf{Sturm}) & None\\
    \hline
        \multicolumn{2}{c|}{\screwnetname (\textbf{ScrewNet})} & Two parts segmented\\
    \end{tabular}
    \caption{Overview of the three methods we evaluate in the full pipeline experiment. \screwnetname is an end-to-end method that handles both visual perception and estimation but requires part segmentations as input.}
    \label{tab:full_pipeline:methods}
\end{table}

In this experiment, we evaluate the category-independent part pose tracking module (\textbf{PT}) proposed in Sec.~\ref{subsec:methodology:category_independent_detection} used in conjunction with the factor graph (\textbf{PT+FG}) against two baselines: \textbf{PT} with \sturmauthor (\textbf{PT+Sturm}) and \textbf{ScrewNet}~\cite{jain_screwnet_2021} (see Table~\ref{tab:full_pipeline:methods}). 

\subsubsection{Setup}

We use simulated data generated from the \mbox{PartNet-Mobility} dataset \cite{xiang_sapien_2020}. Since \textbf{ScrewNet} requires object part segmentations, we provide ground truth segmentations from the simulation. The segmentations are not given to \textbf{PT+FG} or \textbf{PT+Sturm}, since our part tracking module can detect and track object parts without segmentations.

We also modified \textbf{ScrewNet} to predict the twist in the camera frame rather than an object-centric frame (like \cite{DBLP:conf/corl/JainGLN21}).
Additionally, we evaluate \textbf{ScrewNet}'s predictions using our metric introduced in Sec.~\ref{sec:tangent_sim_metric}. 
\begin{appendixcond}
For further details see Appx.~\ref{appendix:sec:metric_screwnet}.
\end{appendixcond}

We conduct three category-independent experiments: 1) train and test on prismatic objects (\texttt{Prismatic}), 2) train and test on revolute objects (\texttt{Revolute}), and 3) train and test on both prismatic and revolute objects (\texttt{Mixed}). To evaluate ability of the methods to generalize across categories, the test sets include only objects from categories not present in the training set (\texttt{Refrigerator} and \texttt{Table}). 
\begin{appendixcond}
For further details, see Appx.~\ref{appendix:full_experiment}.
\end{appendixcond}

% For \sturmauthor, we will retrieve part level poses through using the centers of the trajectory as the position. To retrieve the orientation of each pose, we will use an identity transformation for the first time step and subsequently apply the rotation we have through our relative transformations.

\begin{figure}
    \centering
    \includegraphics[width=\columnwidth]{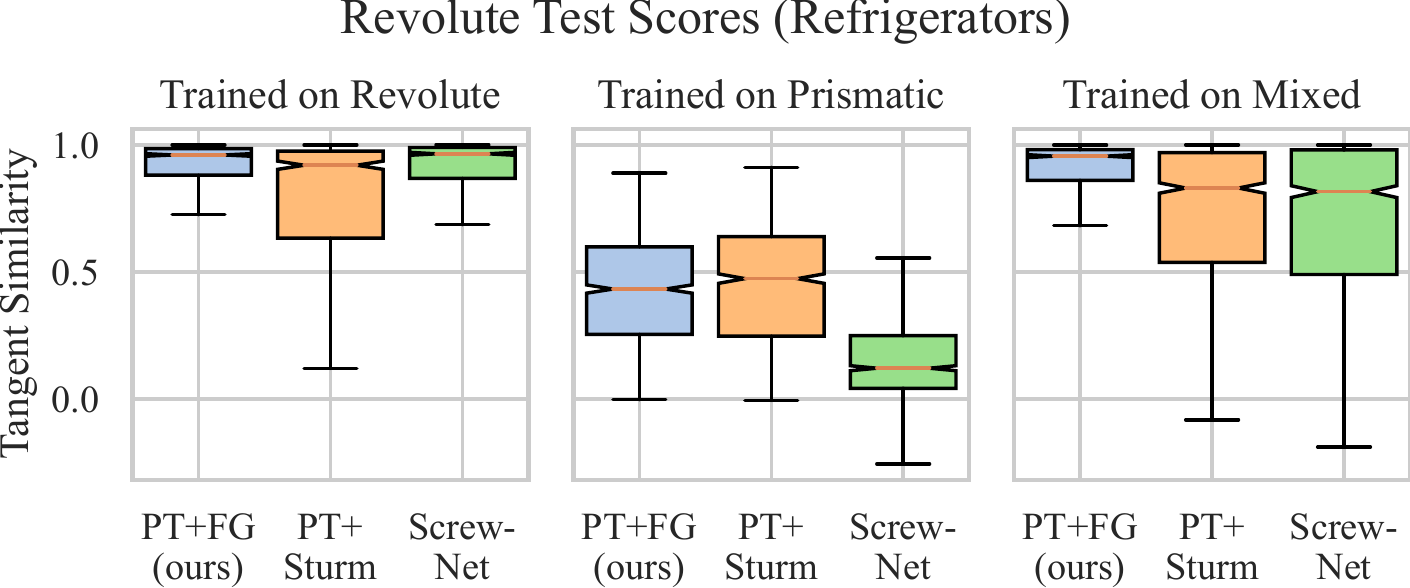}
    
    \vspace{0.4cm}
    
    \includegraphics[width=\columnwidth]{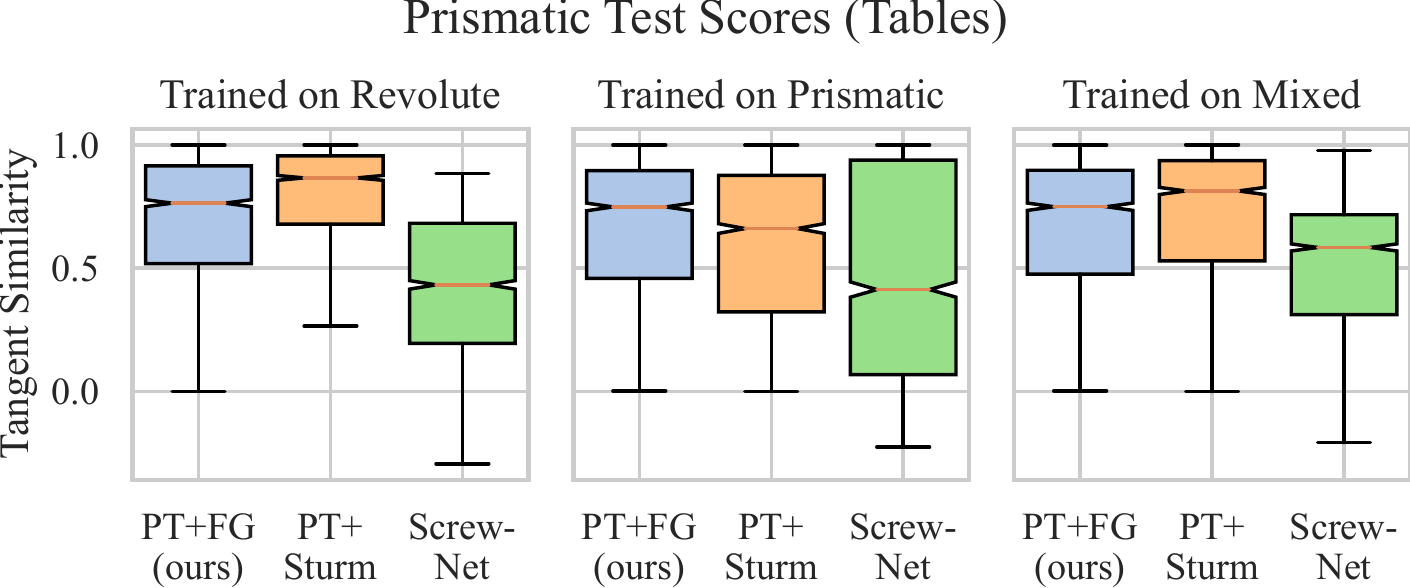}
    % \begin{subfigure}[t]{\linewidth}
    %     \centering
    %     \includegraphics[width=\textwidth]{assets/img/results/full_pipeline/eval_fridges_full.pdf}
    %     \caption{Tested on the unseen \texttt{Fridge} (revolute) category.}
    %     \label{subfig:full_pipeline:fridges}
    % \end{subfigure}%
    % % \hfill
    % \\
    % \begin{subfigure}[t]{\linewidth}
    %     \centering
    %     \includegraphics[width=\textwidth]{assets/img/results/full_pipeline/eval_tables_full.pdf}
    %     \caption{Tested on the unseen \texttt{Table} (prismatic) category.}
    %     \label{subfig:full_pieline:tables}
    % \end{subfigure}
    \caption{Comparison between part tracking+estimation methods (\textbf{PT+FG}, \textbf{PT+Sturm}, \textbf{ScrewNet}) on RGB-D data from PartNet-Mobility. To test category independence, models are trained on \texttt{Prismatic}, \texttt{Revolute}, and \texttt{Mixed} objects, and then evaluated on unseen categories (\texttt{Refrigerator} and \texttt{Table}). Our method (\textbf{PT+FG}) nearly matches or outperforms the baselines in all train/test combinations.}
    \label{fig:full_pipeline_results}
\end{figure}

\subsubsection{Results}

Results for the full pipeline experiment on simulated data are shown in Fig.~\ref{fig:full_pipeline_results}. Our approach (\textbf{PT+FG}) shows the most consistent result across all experiments, while \textbf{PT+Sturm} and \textbf{ScrewNet} are more sensitive to the underlying training data. This demonstrates two points. First, the fact that the underlying training distributions do not matter as much for (\textbf{PT+FG}) indicates that \textbf{PT} is able to learn the concept of articulated part motion on a pixel-level. Second, \textbf{PT} is robust to imbalances in the training set.
%first, our part tracking module learns the general concept of motion on a pixel-level as the underlying training distributions do not seem to matter substantially. And thus, second, the part tracking module can much better deal with the imbalance in the training set. 

\begin{figure}
    \centering
    \includegraphics[width=0.32\linewidth]{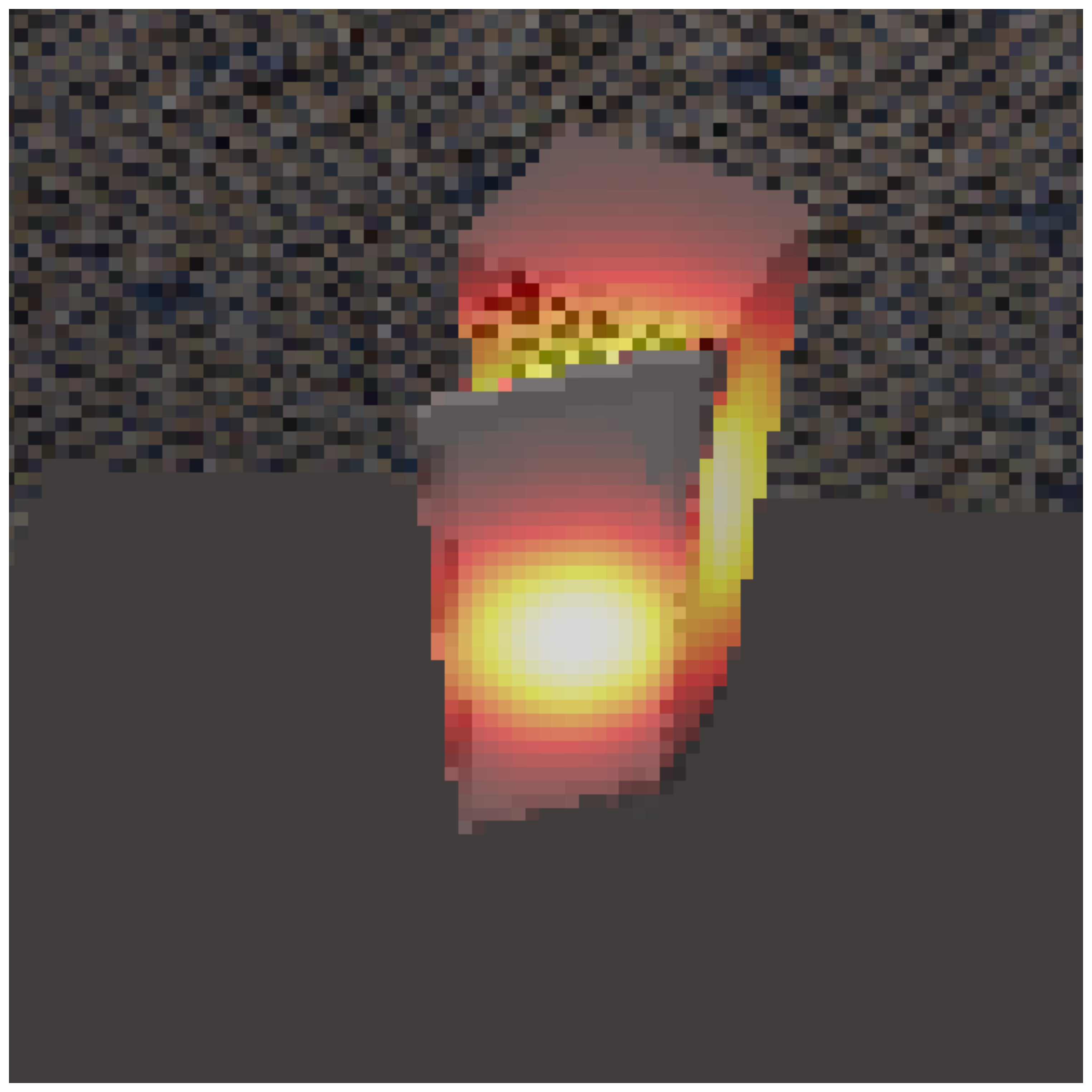}
    \includegraphics[width=0.32\linewidth]{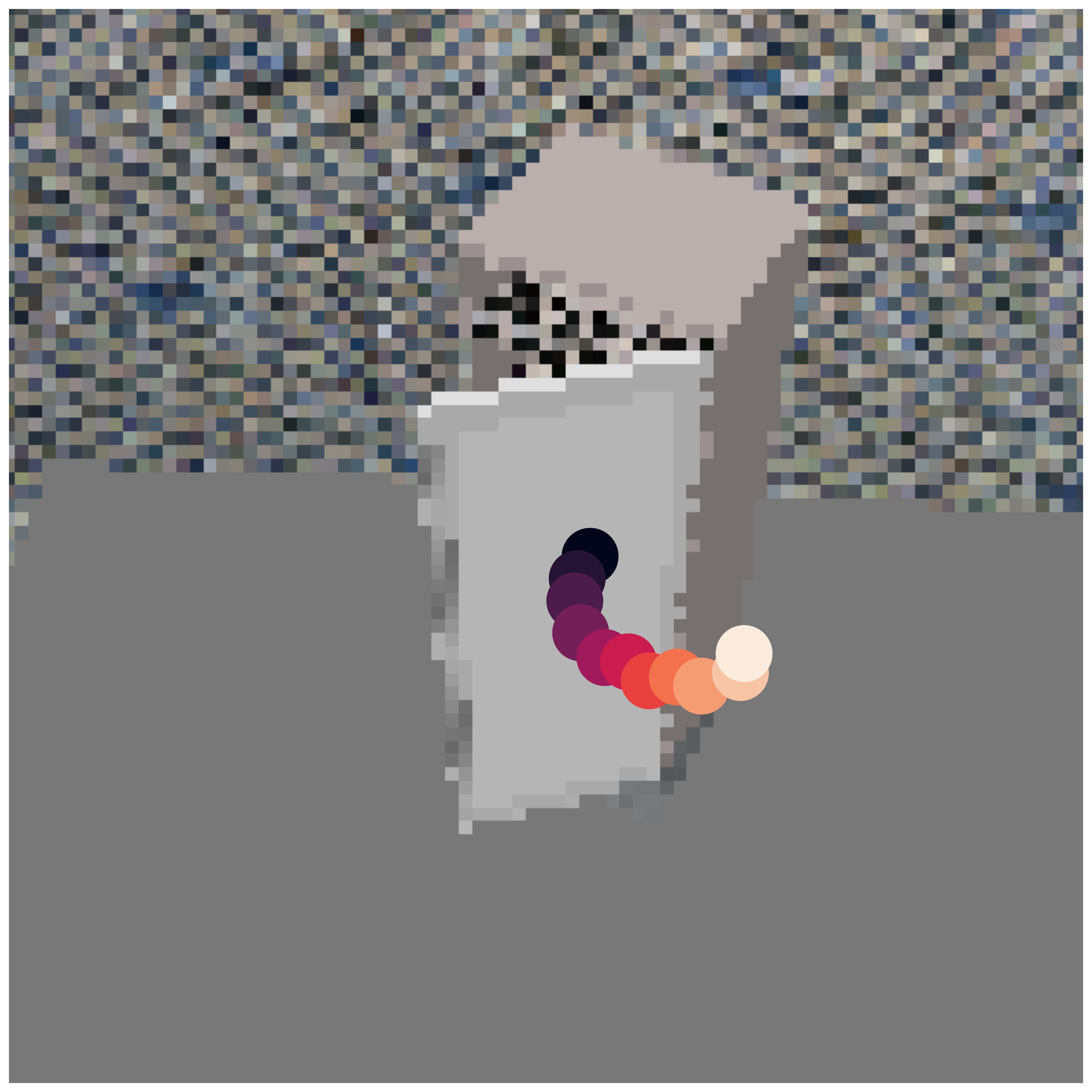}
    \includegraphics[width=0.32\linewidth]{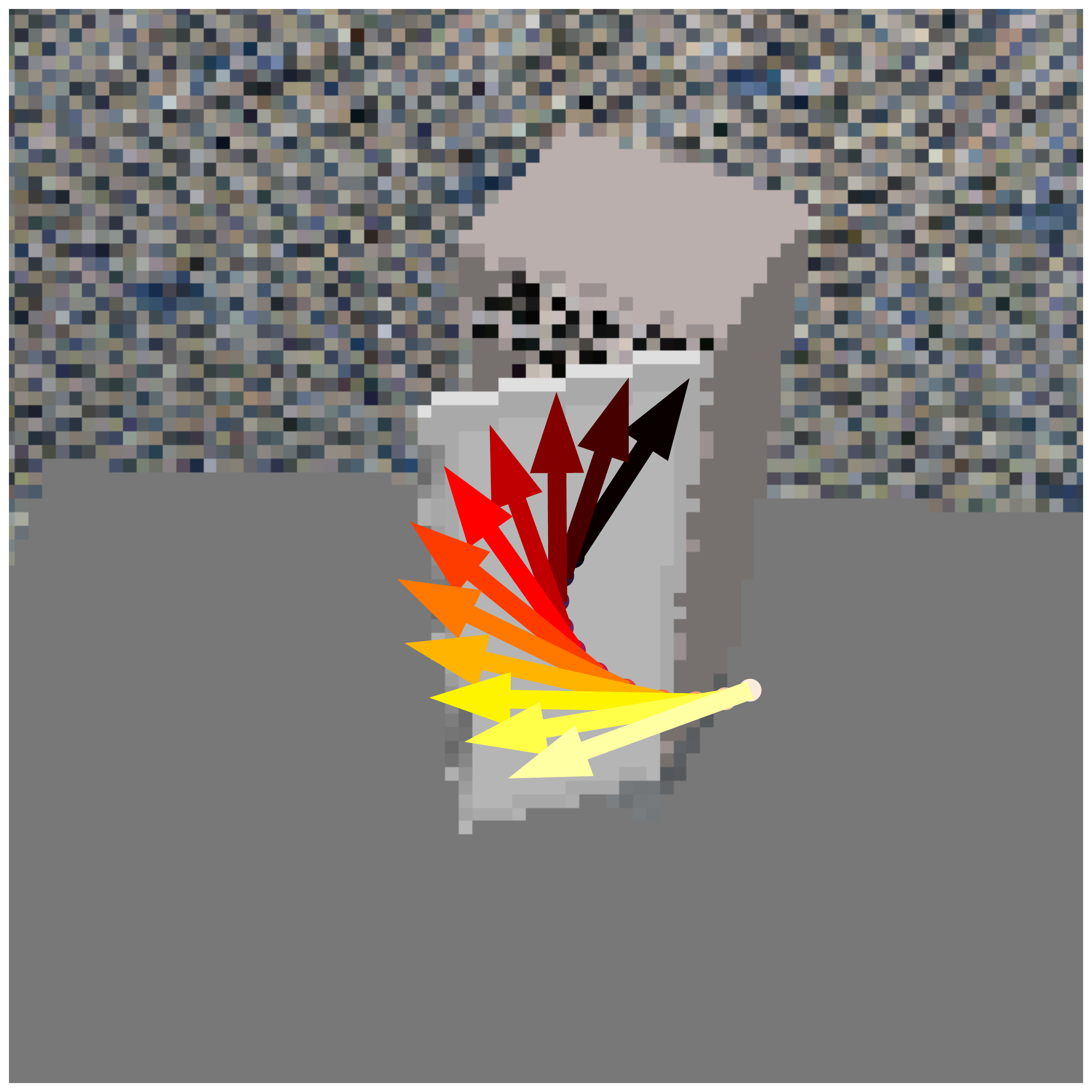}
    \caption{Visualization of our full tracking and estimation pipeline (\textbf{PT+FG}) on a object from the \texttt{Refrigerator} category of the PartNet-Mobility data set. The left image shows the importance predictions output by \textbf{PT}; it learns to focus on the centers of the two cabinet parts (without being given segmentation masks). The middle image shows the predicted part poses output by \textbf{PT}. The right image shows the tangent motion directions of the twist predicted by \textbf{FG}; this prediction achieves a nearly perfect tangent similarity score of 0.999.}
    \label{fig:full_pipeline_qualitative_results}
\end{figure}

Qualitative results are discussed in Fig.~\ref{fig:full_pipeline_qualitative_results}.

%% file: tex/8_conclusion.tex
We present a full pipeline to track and estimate articulated objects on a stream of RGB-D images. We test our estimation method in isolation, outperforming a well established baseline, and show the applicability of our approach on real world data. Our full pipeline performs more consistently than previous category-independent methods without requiring object part segmentation masks. For evaluation, we propose a manipulation-oriented tangent similarity metric that allows a coherent comparison across different joint types.

Possible extensions of this work include applications to articulated objects with multiple joints or leveraging recent techniques for end-to-end optimization through probabilistic state estimators \cite{yi_differentiable_2021, sodhi2022leo, lee_multimodal_2020}. We also plan to extend the tangent similarity metric with a rotation component to capture manipulability for objects like knobs and screw caps.
%Second, removing the two-part assumption and construct a full graph given multiple parts, for example through predicting a binary connectedness score given a pair of part trajectories. And finally, extending our proposed metric that only compares the linear component of the twist to include the rotation component as well, allowing to capture the manipulability for knobs, screw caps, etc.

%\todo{1 col: ends at page 7}

%% file: tex/9_acknowledgements.tex
Toyota Research Institute (TRI) provided funds to assist the authors with their research but this article solely reflects the opinions and conclusions of its authors and not TRI or any other Toyota entity.

%% file: tex/A_screwnet.tex
For each timestep $t$, \textbf{ScrewNet} \cite{jain_screwnet_2021} outputs a tuple $\left(\vec{l}, \vec{m}, \theta, d \right)^{(t)}$ that represents the relative transformation between two parts. Here, $\vec{l}$ and $\vec{m}$ are the Plücker coordinates of the axis $l= \vec{p} + x \vec{l}$, where $\vec{m} = \vec{p} \times \vec{l}$ holds. $\theta$ is the rotation around and $d$ the displacement along the aforementioned axis \cite{jain_screwnet_2021}. The axis is defined in the camera frame.

As derived in \cite{lynch_modern_2017}, we compute the twist $\vec{\nu}^{(t)} = (\vec{v}, \vec{\omega}) \in se(3)$ that describes the relative motion of the body as
\begin{equation}
    \vec{\nu}^{(t)} = 
    \begin{bmatrix}
        \vec{v} \\ \vec{\omega}
   \end{bmatrix} =
   \begin{bmatrix}
       - \theta \vec{m} + d \vec{l} \\
       \theta \vec{l}
   \end{bmatrix}
\end{equation}
given the time-indexed \textbf{ScrewNet} outputs  $\left(\vec{l}, \vec{m}, \theta, d \right)^{(t)}$.

As before, we are interested in the linear velocity along the grasp path. Due to \textbf{ScrewNet}'s time-varying predictions of joint twists $\vec{\nu}^{(t)}$, we need to replace the analytical model introduced in Sec.~\ref{sec:tangent_sim_metric} with a time-indexed approximation. The problem is framed as follows. 

Given both the ground truth twist $\vec{\nu}^{(t)}$ and predicted twist  $\tilde{\vec{\nu}}^{(t)}$ at time $t$, we want to compare their linear velocity at grasping point $\vec{x}(t) \in \mathbb{R}^3$. Similarly to Eq.~\ref{eq:linear_motion}, the linear velocity component at the grasping point is computed as
\begin{align}
     \vec{f} \left(\vec{x}, \vec{\nu} \right)
        &= \vec{\vec{Ad}}_{\lieExp{\vec{x}}^{-1}}(\vec{\nu})_{\vec{v}} \\
        &= \vec{v} + \crossm{\vec{\omega}} \vec{x}.
\end{align}
The time-indexed similarity score then becomes
\begin{align}
    &J \left(
            \vec{\nu}^{(1 : T)}, 
            \tilde{\vec{\nu}}^{(1 : T)}
            \left| \vec{x}^{(1 : T)} \right. 
        \right) \nonumber\\ 
    &\quad= 
        \frac{1}{T} \sum_t 
            \frac{
                \vec{f}
                    \left(
                        \vec{x}^{(t)}, \vec{\nu}^{(t)}
                    \right) \cdot 
                    \vec{f}
                        \left(
                            \vec{x}^{(t)}, \tilde{\vec{\nu}}^{(t)}
                        \right)
            }{
                \|
                    \vec{f}
                        \left(
                            \vec{x}^{(t)}, \vec{\nu}^{(t)}
                        \right)
                \| \| \vec{f}
                        \left(
                            \vec{x}^{(t)}, \tilde{\vec{\nu}}^{(t)}
                        \right) \|
            }.
\end{align}

%% file: tex/C_experiment.tex
\subsection{Factor Graph Experiment (H1)}
\label{appendix:fg_experiment}
\subsubsection{Synthetic Noisy Poses}
% \subsection{Synthetic Noisy Poses Generation}
% \label{appendix:synth_pose_generation}
The goal of this experiment is to create synthetic poses that resemble the most simple kinematic structures of real, articulated household objects that have a fixed base part and one moving articulated part. Initially, the fixed base part $\poselat[][a]^{(t)} \in SE(3)$ is given the same random orientations for all time steps $t \in \left[ 1, \ldots, T \right]$. Then, we sample a sequence of synthetic poses in the following manner. 

First, we sample a random transformation from the first part $a$ to the joint frame $\transform[a][j]$. We uniformly sample an orientation in the full rotation space and a position between $-0.5\si{m}$ and $0.5\si{m}$. We sample an additional random transformation $\transform[j][b]$ from the joint to the second part $b$. The joint state is defined to be zero at the first time step, and thus part $b$ is placed at $\poselat[][b]^{(1)} = \poselat[][a]^{(t)} \transform[a][j] \transform[j][b] \in SE(3)$.

Next, we sample a random canonical twist $\vec{\nu}$ depending on the desired joint type, either prismatic or revolute. Based on that twist, we can then easily generate the full pose sequence for the second part. We denote $q_{\max}$ as the motion magnitude (i.e. how much the joint is actuated). The units of $q_{\max}$ are meters for prismatic joints and radians for revolute joints. We use the notation introduced Sec.~\ref{sec:twist_joint_representation} to represent the joint.
We then define the poses for the second body as
\begin{equation}
    \poselat[][b]^{(t)} = \poselat[][a]^{(t)} \transform[a][j] 
    \lieExp{ \frac{t-1}{T-1} q_{\max} \vec{\nu} }
    \transform[j][b]
\end{equation}
for each time step $t = \left[1, \ldots, T \right]$.

Next, we shift the positions of both bodies such that the overall position mean is 0, centering the full, hypothetical object around the origin.

Lastly, we apply Gaussian noise to all poses to obtain noisy observations for our experiment. The noise is generated by first sampling perturbation twists from a Gaussian distribution
\begin{equation}
    \bm{\nu}_\text{perp} \sim \mathcal{N}(0, \Sigma^2)
\end{equation}
where \mbox{$\Sigma^2 = \text{diag}\left(\sigma_{\text{pos}}^2, \sigma_{\text{pos}}^2, \sigma_{\text{pos}}^2, \sigma_{\text{ori}}^2, \sigma_{\text{ori}}^2, \sigma_{\text{ori}}^2 \right)$}, where position and orientation noise parameters $\sigma_{\text{pos}}^2, \sigma_{\text{pos}}^2$ are defined by the experiment. We apply the perturbation twist by multiplying by its exponentiation
\begin{equation}
    \poseobs[][] = \poselat[][] \lieExp{\bm{\nu}_\text{perp}}.
\end{equation}

\subsection{Full Pipeline Experiment (H2)}
\label{appendix:full_experiment}
\subsubsection{Simulation}
We use a PyBullet Simulation to generate training and test data. A single object from our object instance set (see below) is loaded and randomly translated and rotated around the z-axis before being placed onto a table. Additionally, for each simulation run we randomly sample a camera position and orientation facing the front of the object. We then render RGB-D images for 11 consecutive, different joint configurations. We start with the default joint configuration and randomly increase the joint configuration by a percentage of its max joint range. We randomly sample the percentage from a Gaussian with mean of $8\%$ and standard deviation of $2\%$. Upon reaching the joint limit, we switch the direction of joint motion and start decreasing the joint configuration.

\subsubsection{Object Set}
For our experiments, we consider a set of household categories from the full PartNet-Mobility dataset \cite{xiang_sapien_2020}, namely the categories: \texttt{Box, Dishwasher, Door, Laptop, Microwave, Oven, Refrigerator, StorageFurniture, Table, WashingMachine, Window}. Unlike previous work \cite{jain_screwnet_2021, weng_captra_2021}, we do not manually select instances from the PartNet-Mobility dataset but rather automatically parse all object instances and then filter out instances that have more than one joint, have an unlimited joint range and/or have a joint range less than $10\degree$ or $0.1m$, respectively. An overview of our resulting training and test sets are given in Tab.~\ref{tab:category_independent:training_sets} and Tab.~\ref{tab:category_independent:test_sets}.

\begin{table}[]
    \centering
    \begin{tabular}{c|c|c|c}
        Joint Type & \makecell{Categories} & \makecell{Unique\\instances} & \makecell{Simulation\\Runs} \\
        \hhline{=|=|=|=}
        Revolute & no \texttt{Refrigerator} & 220 & 17516 \\
        \hline
        Prismatic & no \texttt{Table} & 18 &  9311 \\
        \hline
        Mixed & \makecell{no \texttt{Refrigerator}\\and no \texttt{Table}} & $237^*$ & $26748^*$\\
    \end{tabular}
    \caption{Overview of our three training datasets. As the amount of unique instances for the prismatic joint type is low compared to the revolute joint type, we purposely created more simulation runs for these object instances to better balance the data sets.\\
    ${}^*$\textit{The one instance and $79$ simulation runs discrepancy between the sum of the revolute set (first row) and the prismatic set (second row)  to the mixed set (third row) is because the PartNet-Mobility dataset has one \texttt{Table} instance with a revolute joint. This instance was included in the revolute set (first row), and not included in the mixed set (third row).}}
    \label{tab:category_independent:training_sets}
\end{table}

\begin{table}[]
    \centering
    \begin{tabular}{c|c|c|c}
        Joint Type & Categories & Unique instances & Simulation Runs \\
        \hhline{=|=|=|=}
        Revolute & \verb|Fridges| & 14 & 1095 \\
        \hline
        Prismatic & \verb|Tables| & 25 &  1951 \\
    \end{tabular}
    \caption{Overview of our test data sets. }
    \label{tab:category_independent:test_sets}
\end{table}

\subsubsection{Category-Level Experiment}
For completeness, we additionally perform a category-level experiment for our two previously held-out test categories, \texttt{Table} and \texttt{Fridge}. In a category-level experiment, all instances come from the same PartNet-Mobility category, but the training and test set do not share any instances. We split the \texttt{Fridge} instances into 10 training and 4 test instances, and the \texttt{Table} instances into 18 training and 7 test instances. 

\begin{figure}
        \centering
        \includegraphics[width=.7\linewidth]{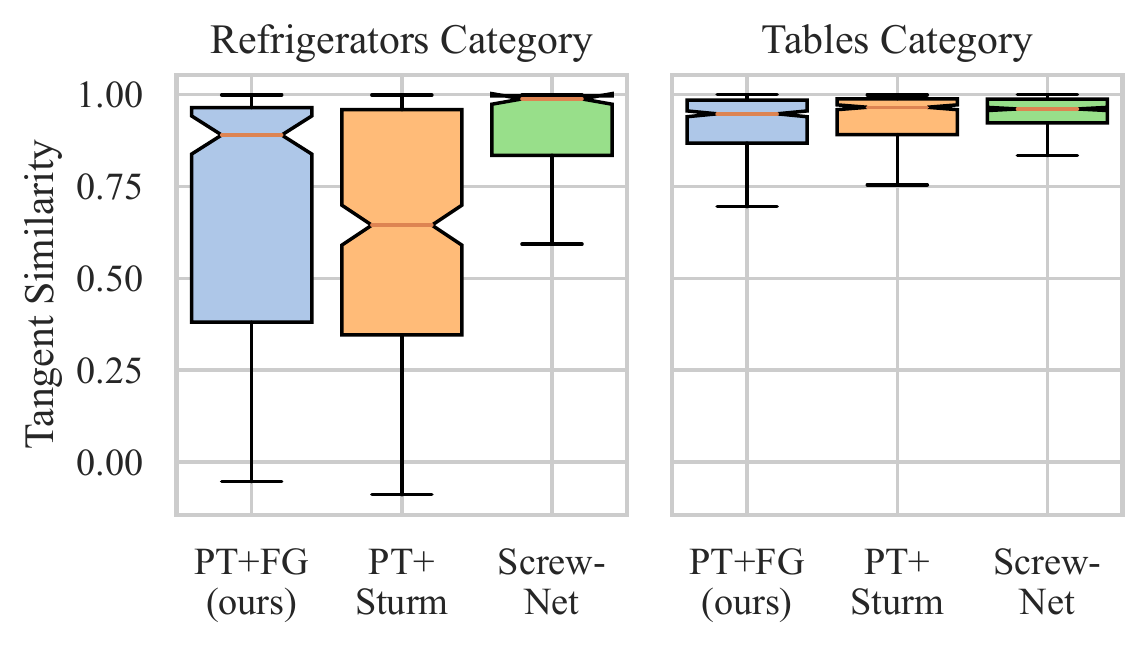}
        \caption{Category level experiment. The worse performance of our method on the \texttt{Fridge} category is due to observed overfitting of our motion feature predictor.}
        \label{fig:full_pieline:category_level}
\end{figure}

Results for this experiment are shown in Fig.~\ref{fig:full_pieline:category_level}. All methods perform well on the \texttt{Table} category but perform worse on the \texttt{Fridge} category. We suspect the main reason is that the \texttt{Fridge} category contains fewer training instances. Additionally, the revolute \texttt{Fridge} joint requires tracking the full 6D pose of the moving part, while the prismatic \texttt{Table} joint only requires tracking translation.

Comparing \textbf{PT+FG} and \textbf{PT+Sturm} to \textbf{ScrewNet}, it is visible that \textbf{ScrewNet} performs better. Compared to \textbf{ScrewNet}, our motion feature predictor outputs low-level motion features, for which the joint type does not matter. As mentioned in Appx.~\ref{appendix:sec:metric_screwnet}, \textbf{ScrewNet} predicts the displacement along and rotation around a predicted axis. Therefore, if all object instances in the training data share the same joint type (i.e. are from the same category as in this experiment), either only the displacement or rotation will vary in the training data. Thus, \textbf{ScrewNet} is given an implicit bias, whereas \textbf{PT} predicts low-level geometric features and does not make use of that bias. Therefore, \textbf{ScrewNet} shows better performance in the cateogry-level experiment. Furthermore, this implicit bias can also explain the worse performance of \textbf{ScrewNet} for the category-independent experiment shown in Fig.~\ref{fig:full_pipeline_results}, specifically the transfer between a model trained on \texttt{Prismatic} and tested on \texttt{Revolute} and vice versa. In this setup, the implicit bias hurts \textbf{ScrewNet}, since the training joint type is different from the test joint type, and \textbf{PT+FG} and \textbf{PT+Sturm} perform better.

Overall, these additional experiments further highlight that a more diverse data set helps for learning general low-level motion features that transfer well in a category-independent setting.

%% file: root.bbl
\begin{thebibliography}{10}
\providecommand{\url}[1]{#1}
\csname url@samestyle\endcsname
\providecommand{\newblock}{\relax}
\providecommand{\bibinfo}[2]{#2}
\providecommand{\BIBentrySTDinterwordspacing}{\spaceskip=0pt\relax}
\providecommand{\BIBentryALTinterwordstretchfactor}{4}
\providecommand{\BIBentryALTinterwordspacing}{\spaceskip=\fontdimen2\font plus
\BIBentryALTinterwordstretchfactor\fontdimen3\font minus
  \fontdimen4\font\relax}
\providecommand{\BIBforeignlanguage}[2]{{%
\expandafter\ifx\csname l@#1\endcsname\relax
\typeout{** WARNING: IEEEtran.bst: No hyphenation pattern has been}%
\typeout{** loaded for the language `#1'. Using the pattern for}%
\typeout{** the default language instead.}%
\else
\language=\csname l@#1\endcsname
\fi
#2}}
\providecommand{\BIBdecl}{\relax}
\BIBdecl

\bibitem{pavlasek_parts-based_2020}
J.~Pavlasek, S.~Lewis, K.~Desingh, and O.~C. Jenkins, ``Parts-based articulated
  object localization in clutter using belief propagation,'' in
  \emph{{IEEE/RSJ} Int. Conf. on Intelligent Robots and Systems (IROS)}, 2020,
  pp. 10\,595--10\,602.

\bibitem{martin-martin_coupled_nodate}
R.~Mart{\'\i}n-Mart{\'\i}n and O.~Brock, ``Coupled recursive estimation for
  online interactive perception of articulated objects,'' \emph{The Int.
  Journal of Robotics Research}, 2019.

\bibitem{sturm_probabilistic_2011}
J.~Sturm, C.~Stachniss, and W.~Burgard, ``A probabilistic framework for
  learning kinematic models of articulated objects,'' \emph{J. Artif. Intell.
  Res.}, vol.~41, pp. 477--526, 2011.

\bibitem{jain_learning_2020}
A.~Jain and S.~Niekum, ``Learning hybrid object kinematics for efficient
  hierarchical planning under uncertainty,'' in \emph{{IEEE/RSJ} Int. Conf. on
  Intelligent Robots and Systems (IROS)}, 2020, pp. 5253--5260.

\bibitem{amato_multiview_2020}
A.~F. Daniele, T.~M. Howard, and M.~R. Walter, ``A multiview approach to
  learning articulated motion models,'' in \emph{Int. Symp on Robotics Research
  ({ISRR})}, 2017, pp. 371--386.

\bibitem{abbatematteo_learning_2019}
B.~Abbatematteo, S.~Tellex, and G.~Konidaris, ``Learning to generalize
  kinematic models to novel objects,'' in \emph{{Conf.} on Robot Learning
  (CoRL)}, 2019, pp. 1289--1299.

\bibitem{michel_pose_2015}
F.~Michel, A.~Krull, E.~Brachmann, M.~Y. Yang, S.~Gumhold, and C.~Rother,
  ``Pose estimation of kinematic chain instances via object coordinate
  regression,'' in \emph{Proc. of the British Machine Vision {Conf.}}\hskip 1em
  plus 0.5em minus 0.4em\relax {BMVA} Press, 2015, pp. 181.1--181.11.

\bibitem{li_category-level_2020}
X.~Li, H.~Wang, L.~Yi, L.~J. Guibas, A.~L. Abbott, and S.~Song,
  ``Category-level articulated object pose estimation,'' in \emph{{IEEE/CVF}
  {Conf.} on Computer Vision and Pattern Recognition {(CVPR)}}, 2020, pp.
  3703--3712.

\bibitem{zeng_visual_2020}
V.~Zeng, T.~E. Lee, J.~Liang, and O.~Kroemer, ``Visual identification of
  articulated object parts,'' in \emph{{IEEE/RSJ} Int. Conf. on Intelligent
  Robots and System {(IROS)}}, 2021, pp. 2443--2450.

\bibitem{DBLP:conf/rss/PillaiWT14}
S.~Pillai, M.~R. Walter, and S.~J. Teller, ``Learning articulated motions from
  visual demonstration,'' in \emph{Robotics: Science and Systems {(RSS)}},
  2014.

\bibitem{hausman_active_2015}
K.~Hausman, S.~Niekum, S.~Osentoski, and G.~S. Sukhatme, ``Active articulation
  model estimation through interactive perception,'' in \emph{{IEEE} {Int.}
  {Conf.} on {Robotics} and {Automation} ({ICRA})}, 2015, pp. 3305--3312.

\bibitem{prats_compliant_2010}
M.~Prats, P.~J. Sanz, and A.~P. del Pobil, ``A framework for compliant physical
  interaction,'' \emph{Auton. Robots}, vol.~28, no.~1, pp. 89--111, 2010.

\bibitem{martin_online_2014}
R.~M. Martín and O.~Brock, ``Online interactive perception of articulated
  objects with multi-level recursive estimation based on task-specific
  priors,'' in \emph{{IEEE}/{RSJ} {Int.} {Conf.} on {Intelligent} {Robots} and
  {Systems} {(IROS)}}, 2014, pp. 2494--2501.

\bibitem{jain_screwnet_2021}
A.~Jain, R.~Lioutikov, C.~Chuck, and S.~Niekum, ``{ScrewNet}:
  Category-independent articulation model estimation from depth images using
  screw theory,'' in \emph{{IEEE} Int. Conf. on Robotics and Automation
  {(ICRA)}}, 2021, pp. 13\,670--13\,677.

\bibitem{weng_captra_2021}
Y.~Weng, H.~Wang, Q.~Zhou, Y.~Qin, Y.~Duan, Q.~Fan, B.~Chen, H.~Su, and L.~J.
  Guibas, ``{CAPTRA:} category-level pose tracking for rigid and articulated
  objects from point clouds,'' in \emph{{IEEE/CVF} Int. Conf. on Computer
  Vision {(ICCV)}}, 2021, pp. 13\,189--13\,198.

\bibitem{liu_nothing_2020}
Q.~Liu, W.~Qiu, W.~Wang, G.~D. Hager, and A.~L. Yuille, ``Nothing but geometric
  constraints: {A} model-free method for articulated object pose estimation,''
  \emph{CoRR}, vol. abs/2012.00088, 2020.

\bibitem{DBLP:conf/corl/JainGLN21}
A.~Jain, S.~Giguere, R.~Lioutikov, and S.~Niekum, ``Distributional depth-based
  estimation of object articulation models,'' in \emph{{Conf.} on Robot
  Learning (CoRL)}, 2021, pp. 1611--1621.

\bibitem{yi_deep_2019}
L.~Yi, H.~Huang, D.~Liu, E.~Kalogerakis, H.~Su, and L.~J. Guibas, ``Deep part
  induction from articulated object pairs,'' \emph{{ACM} Trans. Graph.},
  vol.~37, no.~6, p. 209, 2018.

\bibitem{DBLP:conf/cvpr/YewL20}
Z.~J. Yew and G.~H. Lee, ``Rpm-net: Robust point matching using learned
  features,'' in \emph{{IEEE/CVF} {Conf.} on Computer Vision and Pattern
  Recognition, {(CVPR)}}, 2020, pp. 11\,821--11\,830.

\bibitem{wang_shape2motion_2019}
X.~Wang, B.~Zhou, Y.~Shi, X.~Chen, Q.~Zhao, and K.~Xu, ``{Shape2Motion}:
  {Joint} {Analysis} of {Motion} {Parts} and {Attributes} {From} {3D}
  {Shapes},'' in \emph{{IEEE}/{CVF} {Conf.} on {Computer} {Vision} and
  {Pattern} {Recognition} ({CVPR})}, Jun. 2019, pp. 8868--8876.

\bibitem{shao_motion-based_2018}
L.~Shao, P.~Shah, V.~Dwaracherla, and J.~Bohg, ``Motion-based object
  segmentation based on dense {RGB-D} scene flow,'' \emph{{IEEE} Robotics
  Autom. Lett.}, vol.~3, no.~4, pp. 3797--3804, 2018.

\bibitem{huang_multibodysync_2021}
J.~Huang, H.~Wang, T.~Birdal, M.~Sung, F.~Arrigoni, S.~Hu, and L.~J. Guibas,
  ``Multibodysync: Multi-body segmentation and motion estimation via 3d scan
  synchronization,'' in \emph{{IEEE} {Conf.} on Computer Vision and Pattern
  Recognition (CVPR)}, 2021, pp. 7108--7118.

\bibitem{rofer_kineverse_2022}
A.~R{\"{o}}fer, G.~Bartels, W.~Burgard, A.~Valada, and M.~Beetz, ``Kineverse:
  {A} symbolic articulation model framework for model-agnostic mobile
  manipulation,'' \emph{{IEEE} Robotics Autom. Lett. {(RA-L)}}, vol.~7, no.~2,
  pp. 3372--3379, 2022.

\bibitem{xu2022umpnet}
Z.~Xu, Z.~He, and S.~Song, ``Universal manipulation policy network for
  articulated objects,'' \emph{IEEE Robotics and Automation Letters {(RA-L)}},
  vol.~7, no.~2, pp. 2447--2454, 2022.

\bibitem{he_deep_2016}
K.~He, X.~Zhang, S.~Ren, and J.~Sun, ``Deep residual learning for image
  recognition,'' in \emph{{IEEE/CVF} {Conf.} on Computer Vision and Pattern
  Recognition (CVPR)}, 2016, pp. 770--778.

\bibitem{sodhi_learningtactile_2021}
P.~Sodhi, M.~Kaess, M.~Mukadam, and S.~Anderson, ``Learning tactile models for
  factor graph-based estimation,'' in \emph{{IEEE} Int. Conf. on Robotics and
  Automation {(ICRA)}}, 2021, pp. 13\,686--13\,692.

\bibitem{yi_differentiable_2021}
B.~Yi, M.~Lee, A.~Kloss, R.~Mart\'in-Mart\'in, and J.~Bohg, ``Differentiable
  factor graph optimization for learning smoothers,'' in \emph{IEEE/RSJ Int.
  Conf. on Intelligent Robots and Systems {(IROS)}}, 2021, pp. 1339--1345.

\bibitem{Gill81}
P.~E. Gill, W.~Murray, and M.~H. Wright, \emph{Practical optimization}.\hskip
  1em plus 0.5em minus 0.4em\relax London: Academic Press Inc. [Harcourt Brace
  Jovanovich Publishers], 1981.

\bibitem{xiang_sapien_2020}
F.~Xiang, Y.~Qin, K.~Mo, Y.~Xia, H.~Zhu, F.~Liu, M.~Liu, H.~Jiang, Y.~Yuan,
  H.~Wang, L.~Yi, A.~X. Chang, L.~J. Guibas, and H.~Su, ``{SAPIEN:} {A}
  simulated part-based interactive environment,'' in \emph{{IEEE/CVF} Computer
  Vision and Pattern Recognition (CVPR)}, 2020, pp. 11\,094--11\,104.

\bibitem{sodhi2022leo}
P.~Sodhi, E.~Dexheimer, M.~Mukadam, S.~Anderson, and M.~Kaess, ``Leo: Learning
  energy-based models in factor graph optimization,'' in \emph{{Conf.} on Robot
  Learning (CoRL)}, 2022, pp. 234--244.

\bibitem{lee_multimodal_2020}
M.~A. Lee, B.~Yi, R.~Mart{\'{\i}}n{-}Mart{\'{\i}}n, S.~Savarese, and J.~Bohg,
  ``Multimodal sensor fusion with differentiable filters,'' in \emph{{IEEE/RSJ}
  Int. Conf. on Intelligent Robots and Systems {(IROS)}}, 2020, pp.
  10\,444--10\,451.

\bibitem{lynch_modern_2017}
K.~M. Lynch and F.~C. Park, \emph{\BIBforeignlanguage{en}{Modern robotics:
  mechanics, planning, and control}}.\hskip 1em plus 0.5em minus 0.4em\relax
  Cambridge, UK: Cambridge University Press, 2017, oCLC: ocn983881868.

\end{thebibliography}
